\PassOptionsToPackage{usenames}{color}
\pdfoutput=1 
\documentclass[11pt,letterpaper]{article}
\usepackage{comment}
\usepackage{relsize} 

\usepackage[round]{natbib}
\usepackage{qtree}
\usepackage{imakeidx}  
\makeindex 
\makeindex[name=construals,title={Index of Construals by Scene Role}]
\makeindex[name=revconstruals,title={Index of Construals by Function}]

\usepackage[boxed]{algorithm2e}

\usepackage[small,bf,skip=5pt]{caption}
\usepackage{sidecap} 
\usepackage{rotating}	

\makeatletter
\renewcommand{\paragraph}{%
  \@startsection{paragraph}{4}%
  {\z@}{3.25ex \@plus 1ex \@minus .2ex}{-1em}
  {\normalfont\normalsize\bfseries}%
}
\makeatother

\usepackage[nobottomtitles*]{titlesec}
\titleformat*{\subparagraph}{\itshape}
\titlespacing{\subparagraph}{%
  1em}{
  0pt}{
  1em}

\titleformat{\section}[block]
  {\Large\bfseries}
  {}
  {0pt}
  {\hspace{-1.2cm}
   \makebox[1cm][r]{\normalfont\thesection}\hspace{.2cm}}
\titleformat{name=\section,numberless}[block] 
  {\Large\bfseries}
  {}
  {0pt}
  {\hspace{-1.2cm}
   \makebox[1cm][r]{\normalfont}\hspace{.2cm}}
\titleformat{\subsection}[block]
  {\large\bfseries}
  {}
  {0pt}
  {\hspace{-1.2cm}
   \makebox[1cm][r]{\normalfont\thesubsection}\hspace{.2cm}}
\titleformat{\subsubsection}[block]
  {\normalsize\bfseries}
  {}
  {0pt}
  {\hspace{-1.2cm}
   \makebox[1cm][r]{\normalfont\thesubsubsection}\hspace{.2cm}}

\usepackage{enumitem} 
\setitemize{noitemsep,topsep=0em} 
\setenumerate{noitemsep,leftmargin=0em,itemindent=13pt,topsep=0em}

\usepackage{adjustbox}
\newcommand{\choices}[1]{\adjustbox{stack=ct}{#1}}  

\usepackage{textcomp}

\usepackage[procnames]{listings}

\usepackage{amssymb}	
\usepackage{amsmath}

\usepackage{pifont}

\usepackage{fourier}
\usepackage[scaled=.87]{helvet}
\usepackage[scaled=.8]{beramono}
\usepackage[utf8x]{inputenc}

\usepackage{MnSymbol}	

\usepackage{latexsym}

\usepackage{array}
\usepackage{multirow}
\usepackage{booktabs} 
\usepackage{multicol}
\usepackage{footnote}

\usepackage{hyperref}
\usepackage{url}
\usepackage[usenames]{color}
\usepackage{xcolor}

\definecolor{darkblue}{rgb}{0, 0, 0.5}
\hypersetup{colorlinks=true,citecolor=darkblue, linkcolor=., urlcolor=darkblue}

\usepackage[normalem]{ulem} 
\usepackage{colortbl}
\usepackage{graphicx}
\usepackage{subcaption}
\usepackage{mdframed}

\usepackage{tikz}
\usepackage[edges]{forest}
\usetikzlibrary{arrows,positioning,calc}

\usepackage{color}
\usepackage{bm}
\definecolor{orange}{rgb}{1,0.5,0}
\definecolor{mdred}{rgb}{0.8,0,0}
\definecolor{mdgreen}{rgb}{0,0.6,0}
\definecolor{mdblue}{rgb}{0,0,0.7}
\definecolor{dkblue}{rgb}{0,0,0.5}
\definecolor{dkgray}{rgb}{0.3,0.3,0.3}
\definecolor{slate}{rgb}{0.25,0.25,0.4}
\definecolor{gray}{rgb}{0.5,0.5,0.5}
\definecolor{ltgray}{rgb}{0.7,0.7,0.7}
\definecolor{ltltgray}{rgb}{0.9,0.9,0.9}
\definecolor{purple}{rgb}{0.7,0,1.0}
\definecolor{lavender}{rgb}{0.65,0.55,1.0}

\makeatletter
\lst@AddToHook{EveryPar}{%
  \label{lst:\thelstnumber}
}
\makeatother
\usepackage{cleveref}

\crefformat{part}{\S#2#1#3}
\crefformat{chapter}{\S#2#1#3}
\crefformat{section}{\S#2#1#3}
\crefformat{subsection}{\S#2#1#3}
\crefformat{subsubsection}{\S#2#1#3}
\crefformat{paragraph}{\P#2#1#3}
\crefformat{subparagraph}{\P#2#1#3}
\crefmultiformat{section}{\S#2#1#3}{ and~\S#2#1#3}{, \S#2#1#3}{, and~\S#2#1#3}
\crefmultiformat{subsection}{\S#2#1#3}{ and~\S#2#1#3}{, \S#2#1#3}{, and~\S#2#1#3}
\crefmultiformat{subsubsection}{\S#2#1#3}{ and~\S#2#1#3}{, \S#2#1#3}{, and~\S#2#1#3}
\crefmultiformat{paragraph}{\P\P#2#1#3}{ and~#2#1#3}{, #2#1#3}{, and~#2#1#3}
\crefmultiformat{subparagraph}{\P\P#2#1#3}{ and~#2#1#3}{, #2#1#3}{, and~#2#1#3}
\crefrangeformat{section}{\mbox{\S\S#3#1#4--#5#2#6}}
\crefrangeformat{subsection}{\mbox{\S\S#3#1#4--#5#2#6}}
\crefrangeformat{subsubsection}{\mbox{\S\S#3#1#4--#5#2#6}}
\crefrangeformat{paragraph}{\mbox{\P\P#3#1#4--#5#2#6}}
\crefrangeformat{subparagraph}{\mbox{\P\P#3#1#4--#5#2#6}}
\crefname{part}{Part}{Parts}
\Crefname{part}{Part}{Parts}
\crefname{chapter}{ch.}{ch.}
\Crefname{chapter}{Ch.}{Ch.}
\crefname{figure}{figure}{figures}
\crefname{subfigure}{figure}{figures}
\Crefname{subfigure}{Figure}{Figures}
\crefname{appsec}{appendix}{appendices}
\Crefname{appsec}{Appendix}{Appendices}
\crefname{algocf}{algorithm}{algorithms}
\Crefname{algocf}{Algorithm}{Algorithms}
\crefname{enums}{example}{examples}
\Crefname{enums}{Example}{Examples}
\crefname{enumsi}{example}{examples}
\Crefname{enumsi}{Example}{Examples}
\crefname{}{example}{examples} 
\Crefname{}{Example}{Examples}
\crefformat{enums}{(#2#1#3)}
\crefformat{enumsi}{(#2#1#3)}
\crefformat{}{(#2#1#3)}
\crefname{xnumi}{example}{examples} 
\crefname{xnumi}{example}{examples} 
\Crefname{xnumii}{Example}{Examples} 
\Crefname{xnumii}{Example}{Examples} 
\crefformat{xnumi}{(#2#1#3)} 
\crefformat{xnumii}{(#2#1#3)} 
\crefrangeformat{enums}{\mbox{(#3#1#4--#5#2#6)}}
\crefrangeformat{enumsi}{\mbox{(#3#1#4--#5#2#6)}}
\crefrangeformat{xnumi}{\mbox{(#3#1#4--#5#2#6)}} 
\crefrangeformat{xnumii}{\mbox{(#3#1#4--#5#2#6)}} 
\crefmultiformat{enumsi}{(#2#1#3}{, #2#1#3)}{, #2#1#3}{, #2#1#3)}
\crefmultiformat{xnumi}{(#2#1#3}{, #2#1#3)}{, #2#1#3}{, #2#1#3)} 
\crefmultiformat{xnumii}{(#2#1#3}{, #2#1#3)}{, #2#1#3}{, #2#1#3)} 
\crefrangemultiformat{enumsi}{(#3#1#4--#5#2#6}{, #3#1#4--#5#2#6)}{, #3#1#4--#5#2#6}{, #3#1#4--#5#2#6)}
\crefrangemultiformat{xnumi}{(#3#1#4--#5#2#6}{, #3#1#4--#5#2#6)}{, #3#1#4--#5#2#6}{, #3#1#4--#5#2#6)} 
\crefrangemultiformat{xnumii}{(#3#1#4--#5#2#6}{, #3#1#4--#5#2#6)}{, #3#1#4--#5#2#6}{, #3#1#4--#5#2#6)} 

\ifx\creflastconjunction\undefined%
\newcommand{\creflastconjunction}{, and\nobreakspace} 
\else%
\renewcommand{\creflastconjunction}{, and\nobreakspace} 
\fi%

\newcommand*{\Fullref}[1]{\hyperref[{#1}]{\Cref*{#1}: \nameref*{#1}}}
\newcommand*{\fullref}[1]{\hyperref[{#1}]{\cref*{#1}: \nameref{#1}}}

\usepackage{suffix} 

\usepackage{leipzig}
\usepackage{gb4e} 

\newcommand{\backtick}[0]{\textasciigrave}

\newcommand{\w}[1]{\textit{#1}}	
\newcommand{\p}[1]{\textbf{\textsf{#1}}\index{#1@\textbf{\textsf{#1}}}} 
\newcommand{\punbold}[1]{#1\index{#1@\textbf{\textsf{#1}}}} 
\WithSuffix\newcommand\p*[2]{\textbf{\textsf{#1}}\index{#2@\textbf{\textsf{#2}}}} 
\WithSuffix\newcommand\punbold*[2]{#1\index{#2@\textbf{\textsf{#2}}}} 
\newcommand{\lbl}[1]{\textsc{#1}} 
\newcommand{\psst}[1]{\psstX{#1}{#1}} 
\newcommand{\psstX}[2]{\textcolor{mdgreen}{\hyperref[sec:#1]{\lbl{#2}}}\index{#1@\textcolor{mdgreen}{\textbf{\textsc{#1}}}}} 
\newcommand{\psstdef}[1]{\texorpdfstring{\textcolor{mdgreen}{\hyperref[sec:#1]{\lbl{#1}}}\index{#1@\textcolor{mdgreen}{\textbf{\textsc{#1}}}|textbf}}{#1}} 
\newcommand{\olbl}[1]{\textcolor{purple}{\textrm{#1}}} 

\newcommand{\rf}[2]{\psst{#1}$\leadsto$\psst{#2}\index[construals]{\protect\psst{#1}$\leadsto$\protect\psst{#2}}\index[revconstruals]{#2 #1@\protect\psst{#1}$\leadsto$\protect\psst{#2}}}

\newcommand{\lex}[2][XXXX]{#2\index{#1@#2}} 
\WithSuffix\newcommand\lex*[2][XXXX]{\index{#1@#2}} 

\newcommand{\ghien}[1]{\href{https://github.com/carmls/snacs-guidelines/issues/#1}{\##1en}} 
\newcommand{\ghi}[1]{\href{https://github.com/aryamanarora/carmls-hi/issues/#1}{\##1}} 

\newcommand{\finalversion}[1]{}
\newcommand{\futureversion}[1]{}

\newcommand{\longversion}[1]{} 
\newcommand{\draftnotice}[1]{} 

\newenvironment{discussion}{\begin{mdframed}[linecolor=ltltgray,backgroundcolor=ltltgray]\small\noindent\textit{Discussion.}}{\end{mdframed}}

\newcommand{\hierA}[1]{\textcolor{red}{\hyperref[sec:#1]{#1}}}
\newcommand{\hierB}[1]{\textcolor{blue}{\hyperref[sec:#1]{#1}}}
\newcommand{\hierC}[1]{\textcolor{mdgreen}{\hyperref[sec:#1]{#1}}}
\newcommand{\hierD}[1]{\textcolor{orange}{\hyperref[sec:#1]{#1}}}

\newcommand{\hierAdef}[1]{\section{\psstdef{#1}}\label{sec:#1}}
\newcommand{\hierBdef}[1]{\subsection{\psstdef{#1}}\label{sec:#1}}
\newcommand{\hierCdef}[1]{\subsubsection{\psstdef{#1}}\label{sec:#1}}
\newcommand{\hierDdef}[1]{\paragraph{\psstdef{#1}}\label{sec:#1}}

\newleipzig{cont}{cont}{continuous}
\newleipzig{hab}{hab}{habitual}
\newleipzig{perl}{perl}{perlative}
\newleipzig{emph}{emph}{emphatic}

\title{Hindi--Urdu Adposition and Case Supersenses v1.0}

\newcommand{\emldisplay}[2]{\texttt{\href{mailto:#1}{#2}}}
\newcommand{\eml}[1]{\textsmaller[1.5]{\emldisplay{#1}{#1}}}

\author{
\textbf{Aryaman Arora} \\
  Georgetown University \\
    {\eml{aa2190@georgetown.edu}}\and
\textbf{Nitin Venkateswaran} \\
    Georgetown University \\
    {\eml{nv214@georgetown.edu}} \and
\textbf{Nathan Schneider} \\
    Georgetown University \\
    {\eml{nathan.schneider@georgetown.edu}}
}

\date{\today}

\begin{document}
\maketitle
\begin{abstract}
\noindent
These are the guidelines for the application of SNACS (Semantic Network of Adposition and Case Supersenses; \citealt{schneider-18}) to Modern Standard Hindi of Delhi. SNACS is an inventory of 50 supersenses (semantic labels) for labelling the use of adpositions and case markers with respect to both lexical-semantic function and relation to the underlying context. The English guidelines \citep{en} were used as a model for this document.

Besides the case system, Hindi has an extremely rich adpositional system built on the oblique genitive, with productive incorporation of loanwords even in present-day Hinglish.

This document is aligned with version 2.5 of the English guidelines.
\end{abstract}

\tableofcontents

\newpage

\section{Overview}

This document is \textit{supplementary} to the SNACS v2.5 guidelines for English \citep{en}. It focusses on phenomena specific to Hindi--Urdu, while also attempting to give illustrative examples of the whole inventory of supersenses. We hope this will be useful in annotating typologically similar languages of South Asia, as well as a contribution to the literature on case in Hindi--Urdu.

Taking a page from the Korean guidelines \citep{korean}, we also cover a new top-level supersense group \psst{Context}.

\subsection{Hindi and Urdu}

Hindi and Urdu are two Indo-Aryan-family lects that share a nearly identical grammar, and are best characterised as two diverging registers of one pluricentric language \citep{hindiurdu}. The combined language is generally called Hindi--Urdu or Hindustani in linguistic literature. While the corpus that was annotated during the creation of these guidelines was written in literary Hindi in the Devanagari script, this document aims to cover both Hindi and Urdu.

To that end, all examples are given in transliteration using a system inspired by the International Alphabet of Sanskrit Transliteration (IAST), similar to the rule-based transliteration algorithm used on the English Wiktionary.

Hindi and Urdu diverge lexically even in postposition choice, especially in formal or literary contexts. For example, for the \psst{Locus} postposition meaning `around', Hindi generally uses \p{kī\_cāroṁ\_or} (`on all four sides'; \textit{or} `side' < Sanskrit \textit{avarā}) while Urdu uses \p{ke\_ird-gird} (< Persian \textit{gird} `round'). An attempt is made to give examples from both registers.

\subsection{What counts as an adposition in Hindi--Urdu?}

Following \citet{masica1993indo}, we annotated the Layer II and III function markers in Hindi. These include all of the simple case markers\footnote{\textit{ne} (ergative), \textit{ko} (dative-accusative), \textit{se} (instrumental-ablative-comitative), \textit{k\={a}}/\textit{ke}/\textit{k} (genitive), \textit{me\d{m}} (locative-IN), \textit{tak} (allative), \textit{par} (locative-ON). Declined forms of the pronouns (including the reflexive \textit{apn\={a}}) were also included.} and all of the adpositions.\footnote{An open class, given the productivity of the oblique genitive \textit{ke} as a postposition former.} Our guidelines on the differentially-marked ergative and accusative cases are also applicable to unmarked verbal arguments, but these were not annotated in the first corpus.

We also decided to annotate the suffix \textit{v\={a}l\={a}} when used in an adjectival sense (e.g.~\textit{choṭā-vālā kamrā} `the room that is small'), the comparison terms \textit{jais\={a}} and \textit{jaise}, the extent and similarity particle \textit{sā} (\textit{choṭā-sā kamrā} `small-ish room'), and the emphatic particles \textit{bh}, \textit{hī}, \textit{to} \cite[137--156]{koul}. All of these modify the preceding token and mediate a semantic relation between their object and the object's governor, just as conventionally-designated postpositions do.

\subsection{Background}

This section covers some of the literature and past work we broadly relied on in constructing these guidelines. The main Hindi grammar we referenced was \citet{koul}.

\paragraph{SNACS.} There has been a great deal of work on SNACS across many languages. Those there were generally relevant to this whole document are \citet{schneider-18,en}. For annotating verbal arguments, we started with \citet{shalev-19} which established a baseline for dealing with subjects and objects. Archna Bhatia did some initial work on annotating \textit{The Little Prince} in Hindi in a much earlier SNACS standard.

Comparisons with Korean \citep{korean,hwang-etal-2020-k}, German \citep{german}, and Gujarati\footnote{Personal communication with Maitrey Mehta.} were especially useful in formulating these guidelines. Discussions with the CARMLS research group (particularly Jena Hwang and Vivek Srikumar) and reviewer comments on our work at SIGTYP and SCiL \citep{arora-etal-2020-snacs,arora-etal-2021-snacs} were also instrumental for this work. 

\paragraph{Spatial expressions and motion.} Making sense of the locative cases and their roles as verbal arguments has relied largely on \citet{tafseer} (to disentangle the various functions of locatives) and \citet{narasimhan2003motion} (to understand the framing of motion events).

\paragraph{Verbal arguments.} Much of the guidelines on annotating \psst{Participant}-type roles deal with verbal argument structure. There is a great deal of work on this issue in both linguistics and computational linguistics for Hindi. In theoretical linguistics, there is \citet{mohanan1994argument}, \citet{buttthesis}.

Work on case in Hindi includes general work on differential argument-marking \citep{dehoop2005dam}, dative subjects \citep{dativesubjects,experiencersubjects}, and typology \citep{tafseer}.

The Hindi--Urdu Treebank Project has dominated work on verbal argument structure in computational linguistic work on Hindi. It utilises two models of Hindi syntax: a dependency grammar inspired by the traditional \textit{kāraka} system \citep{vaidya-etal-2011-analysis}, and a modern phrase-structure grammar \citep{hindisyntax,psg}. Bhatt says that the two annotations are analogous to Lexical-Functional Grammar (LFG)'s f-structure and c-structure (when traces are removed from the PSG parse).

Other projects in this field are the Hindi--Urdu PropBank\footnote{The frameset files are available at \url{http://verbs.colorado.edu/propbank/framesets-hindi/}.} \citep{propbank,vaidya-etal-2013-semantic}, the separate Urdu PropBank \citep{anwar-etal-2016-proposition,bhat-etal-2014-adapting}, and Urdu/Hindi VerbNet\footnote{Urdu/Hindi VerbNet took a more SNACS-like approach to annotating lexical semantics of verbal arguments, but was not pursued to make a large resource for verb frames.} \citep{hautli-janisz-etal-2015-encoding}.

\paragraph{Force dynamics.} Some of the biggest issues in porting SNACS to Hindi have been in the realm of force dynamics. Constructions with modal auxiliaries, causatives \citep{begum-sharma-2010-preliminary}, and forced actors are still issues in the guidelines. These are common constructions in South Asian languages, so a resolution to these issues will be necessary as annotation work moves ahead on other languages (e.g.~Gujarati).

\subsection{Organisation}

All examples are written in transliterated form using the International Alphabet of Sanskrit Transliteration (IAST), approximating the spoken pronunciation (i.e. schwa deletion is accounted for). We provide glosses and translations only for illustrative examples in an effort to keep the document concise.

The structure of \psst{Circumstance} and \psst{Configuration} is the same as the English guidelines. For \psst{Participant}, each subsection is a case marker or postposition (instead of a supersense) given the varied functions and scene roles taken on by each marker.

For reference, below is a supersense index for \psst{Participant}. Note that the genitive marker \p{kā} (\cref{sec:genitive}) can nominalise many of these relations.

\begin{itemize}
    \item \psst{Causer}\label{sec:Causer}: \p{ne} (\cref{sec:ergative})
    \item \psst{Agent}\label{sec:Agent}: \p{ne} (\cref{sec:ergative}), \p{ko} (\cref{sec:dative}), \p{se} (\cref{sec:passivesubject})
    \item \psst{Theme}\label{sec:Theme}: \p{ko}, \p{kā}, \p{par} (\cref{sec:accusative})
    \item \psst{Topic}\label{sec:Topic}: \p{ke\_bāre\_meṁ} etc.~(\cref{sec:topic})
    \item \psst{Ancillary}\label{sec:Ancillary}: \p{se}, \p{ke\_sāth} (\cref{sec:comitative}), \p{ke\_binā} (\cref{sec:without})
    \item \psst{Stimulus}\label{sec:Stimulus}: \p{ko}, \p{kā}, \p{par} (\cref{sec:accusative}), \p{se} (\cref{sec:ablative}, \cref{sec:comitative})
    \item \psst{Experiencer}\label{sec:Experiencer}: \p{ko} (\cref{sec:dative}), \p{ne} (\cref{sec:ergative})
    \item \psst{Originator}\label{sec:Originator}: \p{ne} (\cref{sec:ergative}), \p{se} (\cref{sec:ablative})
    \item \psst{Recipient}\label{sec:Recipient}: \p{ko} (\cref{sec:dative}), \p{ne} (\cref{sec:ergative})
    \item \psst{Cost}\label{sec:Cost}: \p{ke\_liye} (\cref{sec:benefactive})
    \item \psst{Beneficiary}\label{sec:Beneficiary}: \p{ke\_liye} (\cref{sec:benefactive}), \p{ke\_xilāf}, \p{ke\_viruddh} (\cref{sec:against}), \p{ko} (\cref{sec:dative})
    \item \psst{Instrument}\label{sec:Instrument}: \p{se} (\cref{sec:instrumental})
\end{itemize}

\newpage
\hierAdef{Circumstance}

\begin{table}[ht]
    \centering
    \begin{tabular}{rccccccc}
        \toprule
        & \Obl & \p{meṁ} & \p{par}/\p{pe} & \p{se} & \p{tak} & \p{ko} & \p{ke\_lie}\\
        \midrule
        \psst{Circumstance} & & $\checkmark$ & $\checkmark$ & & & & $\checkmark$\\
        \midrule
        \psst{Locus} & & $\checkmark$ & $\checkmark$ & & & &\\
        \psst{Source} & & & & $\checkmark$ & & &\\
        \psst{Goal} & $\checkmark$ & & & & $\checkmark$ & $\checkmark$ &\\
        \psst{Extent} & $\checkmark$ & & & & $\checkmark$ & & $\checkmark$\\
        \midrule
        \psst{Time} & $\checkmark$ & $\checkmark$ & $\checkmark$ & & & $\checkmark$ &\\
        \psst{StartTime} & & & & $\checkmark$ & & &\\
        \psst{EndTime} & & & & & $\checkmark$ & &\\
        \psst{Duration} & $\checkmark$ & $\checkmark$ & & & & & $\checkmark$\\
        \bottomrule
    \end{tabular}
    \caption{Functions for some of the basic spatio-temporal postpositions and case markers. Note that the \psst{Locus} scene is permissible for \p{se} and \p{tak} in fictive motion, and that \p{meṁ} and \p{par}/\p{pe} can take a \psst{Goal} scene when licensed by a motion verb. \psst{Direction} and \psst{Interval} are spatio-temporal parallels, but are not in this table because they are marked by several idiosyncratic postpositions.}
\end{table}

\psst{Circumstance} is used directly as a scene role when some additional information is added to contextualize the main event. These tend to involve locative postpositions: \p{meṁ}, \p{par}, etc.

\begin{exe}
    \ex \gll durghaṭnā \p{meṁ} do log ghāyal hue. \\
             accident {\Loc} two people injured be.\Pfv \\
        \glt `Two people were injured in the accident.'
    \ex \gll Koronā kāl \p{ke\_calte}... (\rf{Circumstance}{Time})\\
             coronavirus epoch during\\
        \glt `As the Age of Coronavirus continued...'
\end{exe}

It is used for \textbf{setting events}, often construed as a \psst{Locus} and perhaps serving as an answer to a location-based question, but the postposition itself does not give an explicit location.

\begin{exe}
    \ex \gll kām \p{par} (\rf{Circumstance}{Locus}) \\
             work {\Loc} \\
        \glt `at work'
\end{exe}

It is also used for \textbf{occasions}, when the event is only the background for the action (rather than a cause).

\begin{exe}
    \ex \gll janamdin \p{ke\_lie} kyā kiyā? \\
             birthday for what do.\Pfv \\
        \glt `What did you do for your birthday?'
    \ex \gll āpne la\~{n}c \p{meṁ} kyā khāyā? \\
             {you.\Erg} lunch {\Loc} what eat.\Pfv \\
        \glt `What did you eat for lunch?'
\end{exe}

\hierBdef{Temporal}

Not used directly so far.

\hierCdef{Time}

\p{meṁ} indicates temporal placement in the context of some span of time (e.g.~a day, a month, a century). \p{ko}, \p{par}, and \p{pe} are optionally used in a similar manner \citep{koul}. Note that these fixed time postpositional markers are often optional.

\begin{exe}
    \ex \begin{xlist}
    \ex \gll ham jānvarī \p{meṁ} mile\.{n}ge.\\
             we January {\Loc} {meet.\Fut} \\
        \glt `We will meet in January.'
    \ex \gll kam umr \p{meṁ} \\
             less age {\Loc} \\
        \glt `at a young age'\end{xlist}
    \ex \gll kaun jāne kal \p{ko} kyā hogā? \\
             who know tomorrow {\Dat} what {be.\Fut} \\
        \glt `Who knows what will happen in the future?'
    \ex 2012 kī ardhrātri \p{ke\_samay} śahar ke logoṁ ko ek dhamāke kī āvāz ne ḍarā diyā.
\end{exe}

Relative time markers such as \p{ke\_bād} ``after'' and \p{se\_pahle} ``before'' are also included. However, if the difference in time is explicitly stated that the construal \rf{Time}{Interval} is used.

\begin{exe}
    \ex \gll disambar \p{ke\_bād} \\
             December after \\
             after December
    \ex \gll disambar \p*{ke\_}{ke\_bād} do mahīne \p*{\_bad}{ke\_bād} (\rf{Time}{Interval})\\ 
             December {\Gen} two months after \\
             two months after December
\end{exe}

Finally, adpositions that pick out an arbitrary point in time from a duration such as \p{ke\_daurān} ``during'', \p{kī\_avdhi\_meṁ} ``in the interval of'' also take this as scene role and function.

\begin{discussion}
This is the only context in which \p{ko} would create an adverb. It doesn't fit under any other function very well. \rf{Time}{Goal} was considered at some point but the grammatical functions are entirely different.

It was elected to not mix time and location in construals, following the precedent of \citet{en}.
\end{discussion}

\hierDdef{StartTime}

The prototypical postposition is \p{se}.

\begin{exe}
    \ex \gll mujhe kal \p{se} ṭhanḍ lag rahī hai.\\
             {I.\Dat} yesterday {\Abl} coldness feel {\Cont} {be.\Prs} \\
        \glt `I have been feeling cold since yesterday.'
\end{exe}

Unlike the equivalent English \textit{since}, \p{se} can also be used to delineate the beginning of an interval of time.\footnote{This difference is especially apparent in Indian English, where even the formal register permits constructions like \textit{since two years} for standard \textit{since two years ago}.} In this sense, it was decided the construal \rf{StartTime}{Interval} is appropriate; the interval is not specified to have an endpoint so it does not fit the definition of \psst{Duration}.\footnote{\cite{goel-etal-2020-hindi}, in Hindi TimeBank, classify this as a \psst{Duration}, on the basis that an interval of time is referred to (regardless of whether the \psst{EndTime} is known).}

\begin{exe}
    \ex \gll barsoṁ \p{se} yuddh ho rahī hai\\
             years {\Abl} war be {\Cont} {be.\Prs} \\
        \glt `The war has been raging since years ago.'
\end{exe}

\hierDdef{EndTime}

The prototypical postposition is \p{tak}. \psst{StartTime} is an exact counterpart of this, and the \rf{EndTime}{Interval} construal applies for a durative use.

\begin{exe}
    \ex \gll kal se kal \p{tak}\\
             yesterday {\Abl} tomorrow {\All}\\
        \glt `from yesterday until tomorrow'
\end{exe}

\begin{discussion}
For the durative uses of \p{se} and \p{tak} it was difficult to come to a consensus on the label; the alternative option (e.g.~for \p{se}) was \rf{Duration}{StartTime}. We felt that the difference between durative and non-durative was morphosyntactic rather than semantic.
\end{discussion}

\hierCdef{Frequency}

The prototypical examples for \psst{Frequency} are expressed through reduplication (e.g.~\textit{kabhī-kabhī}  `sometimes') rather than a postposition.

For iterations marked ordinally with \p{ke\_lie}, \psst{Frequency} is used:

\begin{exe}
    \ex \gll tīsrī bār \p{ke\_lie}\\
             third time for \\
             for the third time
\end{exe}

\hierCdef{Duration}

\psst{Duration} covers two types of postpositions that are distinct in Hindi. \p{meṁ} focuses on the duration involved in achieving some outcome.

\begin{exe}
    \ex \begin{xlist}
        \ex \gll kitne din \p{meṁ} likh pāoge? \\
                 {how.many} days {\Loc} write {be.able.\Fut} \\
            \glt `In how many days will you be able to write it?'
        \ex \gll do sāl \p{meṁ} do bār kiyā. \\
                 two years {\Loc} two times {do.\Pfv} \\
            \glt `I did it twice in two years.'
    \end{xlist}
\end{exe}

\p{ke\_lie} focuses on the duration over which an action occurs. The action occurs continuously over that span.

\begin{exe}
    \ex \begin{xlist}
        \ex \gll kitne din \p{ke\_lie} likh pāoge? \\
                 {how.many} days for write {be.able.\Fut} \\
            \glt `For how many days will you be able to write? [e.g.~said to a journalist]'
    \end{xlist}
\end{exe}

\hierCdef{Interval}

This role is fulfilled by plain \p{pahle} `ago' and \p{bād} `later' when they are attached to a unit of time. Note that by themselves they are adverbs meaning `earlier' and `later'.

\begin{exe}
    \ex \gll do sāl \p{pahle} \\
             two years ago\\
        \glt `two years ago'
\end{exe}

\hierBdef{Locus}

\psst{Locus} is prototypically used to indicate a static location, whether literal or abstract (e.g.~location on the Internet).

For \p{meṁ} `in', this is within some enclosing entity (e.g.~a geographical area, a container, a building). It cannot be a point location. \p{ke\_andar} `inside of' functions similarly.

\begin{exe}
    \ex \begin{xlist}
        \ex \gll maiṁ mumbaī \p{meṁ} rahtā hūṁ.\\
                 {\First\Sg} Mumbai {\Loc} {stay.\Prs.\Hab} {be.\Prs} \\
            \glt `I live in Mumbai.'
        \ex \gll us bakse \p{meṁ} kyā hai?\\
                 that box {\Loc} what {be.\Prs}\\
            \glt `What is in that box?'
    \end{xlist}
    \ex baṛe-baṛe deśoṁ \p{meṁ} aisī choṭī-choṭī bāteṁ hotī rahtī haiṁ, Senyoritā!
    \ex bakse \p{ke\_andar}
\end{exe}

For \p{par} and \p{pe}, on the other hand, the location may be a point, but it has to be an entity on top of or over which something can be placed.

\begin{exe}
    \ex \begin{xlist}
        \ex \gll ghar \p{pe}\\
                 home at\\
            \glt `at home'
        \ex bakse \p{par}
    \end{xlist}
\end{exe}

All the other various relative static location adpositions are treated as \psst{Locus} as well.

\begin{exe}
    \ex zamīn \p{ke\_ūpar}
    \ex āsmān \p{ke\_nīce}
    \ex gāṛī \p{ke\_pās} do log khaṛe haiṁ
    \ex \gll us\p{kī\_cāroṁ\_or} pānī thā\\
             {3\Sg.\Gen.around} water be.\Pst\\
        \glt `There was water all around her.'
\end{exe}

Hindi also can express static locations using dynamic postpositions, a phenomenon called \textbf{fictive motion}.

\begin{exe}
    \ex \gll chat \p{se} pūrā śahar dikhtā hai (\rf{Locus}{Source})\\
             roof {\Abl} full city {be.seen.\Hab} {be.\Prs}\\
        \glt `The whole city is visible from the roof.'
    \ex saṛak nadī \p{tak} jātī hai. (\rf{Locus}{Goal})
\end{exe}

\paragraph{Connection verbs.} Various verbs that indicate connection and take an argument in the comitative, when dealing with static events, are labelled \rf{Locus}{Ancillary}.

\begin{exe}
    \ex \gll nāv peṛ \p{se} bandhī hai.\\
             boat tree {\Com} {be.tied.\Pfv} {\Cop.\Prs}\\
        \glt `The boat is tied to the tree.'
\end{exe}

These also have motion equivalents when licensed by a non-stative verb. See under \psst{Goal}.

\begin{discussion}
It is unclear how to treat habitual tense verbs that can be ambiguously construed as static or dynamic.

\begin{exe}
    \ex \gll nadī samundar \p{tak} bahtī hai.\\
             river ocean {\All} {flow.\Hab} {\Cop.\Prs}\\
        \glt The river flows till the ocean.
\end{exe}

Is this a statement of fact about where the river ends (thus \rf{Locus}{Goal}), or is it the present flowing of the river to that endpoint (thus \psst{Goal})? We fall back on the most literal reading (so \psst{Goal}) in case of ambiguity. This is part of an open issue cross-lingually, see \ghien{120}.

\end{discussion}

\hierCdef{Source}

The prototypical postposition for this is \p{se}, which often takes on the \psst{Source} function even in other roles. In this function it is comparable to English \w{from}.

\begin{exe}
    \ex \gll vah kal hī dillī \p{se} niklī.\\
             {\Third\Sg} yesterday {\Emph} Delhi {\Abl} {leave.\Pfv}\\
        \glt `She left Delhi just yesterday.'
\end{exe}

This scene also covers initial states before a transformation.

\begin{exe}
    \ex \gll maiṁne māṭī \p{se} banāyā.\\
             {\First\Sg.\Erg} clay {\Abl} {make.\Pfv}\\
        \glt `I made it out of clay.'
\end{exe}

\hierCdef{Goal}

\psst{Goal} indicates a final location or state. Many motion verbs do not explicitly mark the endpoint of motion, instead treating it as a direct object. The case markers \p{ko} and \p{tak} do have prototypical \psst{Goal} functions, and can be optionally used to mark those objects.

\begin{exe}
    \ex \gll maiṁ dillī (\p{ko}) gayā.\\
             {\First\Sg} Delhi to go.\Pfv\\
        \glt `I went to Delhi.'
\end{exe}

All of the locative postpositions and case markers can take on a \psst{Goal} scene role if licensed by a motion verb. Hindi syntactically patterns with verb-framed languages, but path is usually lexicalized in postpositions \citep{narasimhan2003motion}.

\paragraph{Connection verbs.} Various verbs that indicate connection and take an argument in the comitative, when dealing with dynamic events, are labelled \rf{Goal}{Ancillary}.

\begin{exe}
    \ex \gll nāv ko peṛ \p{se} bāṁdho.\\
             boat {\Acc} tree {\Com} {tie.\Imp}\\
        \glt `Tie the boat to the tree.'
\end{exe}

These also have static senses. See under \psst{Locus}.

\hierBdef{Path}

This is traditionally called the perlative case, which is expressed with the ubiquitous \p{se}. Unlike English, there is not much variety in \psst{Path} adpositions (\textit{over}, \textit{across}, \textit{through}, as well as uses of static location markers), but postposition stacking is permissible with \p{se}.

\begin{exe}
    \ex \gll railī dillī \p{se} guzrī thī.\\
             rally Delhi {\Perl} {pass.\Pfv} {\Cop.\Pst}\\
        \glt `The rally passed through Delhi.'
    \ex \begin{xlist}
            \ex \gll havāī-jahāz me\punbold*{re\_ūpar}{ke\_ūpar}\textsubscript{\psst{Locus}} \p{se} gayā.\\
                 {airplane} {\First\Sg.above} {via} {go.\Pfv}\\
            \glt `The plane flew over me.'
            \ex vo śaitān me\punbold*{re\_pīche}{ke\_pīche}\textsubscript{\psst{Locus}} \p{se} bhāg gayā!
        \end{xlist}
\end{exe}

\p{se\_hokar} also marks a \psst{Path} \citep[p. 150]{narasimhan2003motion}.

\begin{discussion}
There is a \rf{Path}{Instrument} construal in English for e.g.~``escape \textbf{by} tunnel'', but there does not seem to be anything instrumental about the equivalent Hindi construction, so we just treat it as a \psst{Path}.
\end{discussion}

\hierCdef{Direction}

\psst{Direction} is the static or dynamic orientation of something. The prototypical markers for this are \p{kī\_taraf}, \p{kī\_or}, and \p{kī\_diśā}, all grammaticalised from the literal meaning `in the direction of'.

\begin{exe}
    \ex \gll maiṁ darvāze \p{kī\_taraf} cal rahā thā.\\
             {\First\Sg} {door.\Obl} {in.direction.of} walk {\Cont} {\Cop.\Pst}\\
        \glt `I was walking towards the door.'
\end{exe}

Like in English, some motion adverbs\footnote{Unlike in English, where words like \w{behind}, \w{up}, etc.~can also be analysed as intransitive prepositions or particles, there is generally no disagreement in Hindi grammar on the status of motion adverbs as adverbs. They are formed from the oblique case of nouns, like many other non-motion adverbs.} satisfy the definition of \psst{Direction} and thus are fair game for annotation. A list of these is in \cref{table:motion}.
\begin{exe}
    \ex \gll maiṁ \p{bāhar} jāne kā soc rahā thā.\\
             {\First\Sg} outside {go.\Inf.\Obl} {\Gen} think {\Cont} {\Cop.\Pst}\\
        \glt `I was thinking of going outside.'
    \ex vo \choices{\p{sīdhe}\\\p{dāyeṁ}\\\p{bāyeṁ}\\\p{vāpas}} calā.
\end{exe}

\paragraph{Distance.} Static distance uses the construal \rf{Locus}{Direction}, since it refers to a fixed point in space but in a way as to emphasise the distance is movement away from another point. \p{se\_dūr} and \p{ke\_dūr} are used in this way.

\begin{exe}
    \ex \gll dillī hamāre gāṁv \p*{se\_}{se\_dūr} bīs kilomīṭar \p*{\_dūr}{se\_dūr} hai.\\
             Delhi {\First\Pl.\Gen} town {\Abl} twenty kilometres far {\Cop.\Prs}\\
        \glt `Delhi is ten kilometres away from our town.'
\end{exe}

\begin{table}
    \centering
    \begin{tabular}{|c|c|}
    \hline
        \p{āge} & ahead \\
        \p{sāmne} & in front\\
        \p{ūpar} & up \\ 
        \p{dāyeṁ}, \p{dāhine}, \p{sīdhe} & right \\ 
        \p{sīdhe} & straight \\
        \p{dūr} & far \\ 
    \hline
    \end{tabular}
    \begin{tabular}{|c|c|}
    \hline
        \p{pīche} & behind \\ 
        \p{bāhar} & outside \\
        \p{nīce} & down \\ 
        \p{bāyeṁ}, \p{ulṭe} & left \\ 
        \p{ulṭe} & backwards \\
        \p{pās}, \p{qarīb}, \p{nazdīk} & near \\ 
    \hline
    \end{tabular}
    \caption{Some motion adverbs in Hindi.}
    \label{table:motion}
\end{table}

\hierCdef{Extent}

When referring to scalar values or changes on a scale, \p{tak} has the role of \psst{Extent}. \p{kā} can function similarly, but take a construal \rf{Extent}{Identity} since it equates two things.

\begin{exe}
    \ex \gll hameṁ mīloṁ \p{tak} bhāgnā paṛā.\\
             {\First\Pl.\Dat} {miles.\Obl} {\All} {run.\Inf} {have.to.\Pfv}\\
        \glt `We had to run for miles.'
    \ex \gll sau rupaye \p{kā} munāfā (\rf{Extent}{Identity})\\
             hundred rupees {\Gen} profit\\
        \glt `a profit of 100 rupees'
\end{exe}

Often, this kind of semantic relation is not marked by an adposition or case marker, and is instead a core argument of the verb.

\begin{exe}
    \ex \gll dām [das pratiśat]\textsubscript{\psst{Extent}} baṛhā.\\
             price ten percent {increase.\Pfv}\\
        \glt `The price increased by 10\%.'
\end{exe}

\paragraph{\p{jitnā} ... \p{utnā}} Hindi's \p{jitnā} ... \p{utnā} construction functions exactly the same as the English \w{as ... as} (except reversed) and is annotated on the same semantics.

\begin{exe}
    \ex \gll vo \p{jitnā}\textsubscript{\psst{ComparisonRef}} kar saktā thā \p{utnā}\textsubscript{\psst{Extent}} usne kiyā.\\
             {\Third\Sg} {as.much} do {can.\Hab} {\Cop.\Pst} {that.much} {\Third\Sg.\Erg} {do.\Pfv}\\
        \glt (S)he did as much as (s)he could.
    \ex jagah \p*{jitnī}{jitnā}\textsubscript{\psst{ComparisonRef}} sundar hai \p*{utnī}{utnā}\textsubscript{\rf{Characteristic}{Extent}} xatarnāk hai.
\end{exe}

\paragraph{\p{sā}} The postposition \p{sā} is difficult to translate succinctly into English, but in that sense that it means `rather' or `pretty' (when modifying an adjective) it is best labelled by \psst{Extent}. The other sense of it is covered under \psst{ComparisonRef}.

\begin{exe}
    \ex \gll acchā \p{sā} ādmī\\
             good rather man\\
        \glt `a rather good man'
\end{exe}

\hierBdef{Means}

\psst{Means} describes a secondary action or event utilised towards performing the verb at hand. A gerund (or other nominal that refers to an action) as an instrumental argument marked with \p{se} to a verb is \psst{Means}.

\begin{exe}
    \ex \gll unhoṁne golībārī \p{se} badlā liyā.\\
             {\Third\Pl.\Erg} shooting {\Ins} revenge {take.\Pfv}.\\
        \glt `They retaliated with shootings.'
    \ex zyādā tez bhāgne \p{se} ṭāṁg toṛ dī.
\end{exe}

\hierBdef{Manner}

The \textit{how} of a situation, usually an adverbial phrase.

\begin{exe}
    \ex \begin{xlist}
    \ex \gll merī bāt dhyān \p{se} suno.\\
             {\First\Sg.\Gen} talk care {\Ins} {list.\Imp}\\
        \glt `Listen to me carefully.'
    \ex \choices{ġaltī\\zor\\pyār} \p{se}
        \end{xlist}
    \ex \begin{xlist}
    \ex \gll usne ġusse \p{meṁ} kah diyā.\\
             {\Third\Sg.\Erg} anger {\Loc} say {give.\Pfv}\\
        \glt `He rashly said it in anger.'
    \ex ham gujarātī \p{meṁ} bāt kar rahe haiṁ.
    \end{xlist}
    \ex \gll {agar} {āp} \p*{binā\_}{ke\_binā} {ovan} \p*{\_ke}{ke\_binā} {kek} {banā} {rahī} {hai} \\
    {if} {\Second\Pl} {without} {oven} {\Gen} {cake} {make} {\Cont} {\Cop.\Prs} \\
    \glt `if you are making a cake without an oven'
\end{exe}

When a comparison postposition is used adverbially (e.g.~\p{jaise}) it also gets the scene role of \psst{Manner}.

\begin{exe}
    \ex \rf{Manner}{ComparisonRef}: \begin{xlist}
        \ex \gll tū jānvar \p{kī\_tarah} khātā hai.\\
                 {\Second\Sg} animal {like} {eat.\Hab} {\Cop.\Prs}.\\
            \glt `You eat like an animal.'
        \end{xlist}
\end{exe}

\hierBdef{Explanation}

The \textit{why} of a situation. The instigating event of another event is an \psst{Explanation}.

\begin{exe}
    \ex \begin{xlist}
        \ex \gll me\p*{rī\_vajah\_se}{kī\_vajah\_se} sab gaṛbaṛ huā.\\
                 {\First\Sg.because} everything messed.up {\Cop.\Pfv}\\
            \glt `Everything went wrong because of me.'
        \ex uske na jāne \p{ke\_kāraṇ} maiṁ ghar pe rahā.
    \end{xlist}
    \ex ġusse \p{se} rūṭhnā (\rf{Explanation}{Source})
    \ex \gll aur dambh \p{se} phūlā rahtā hai. (\rf{Explanation}{Source})\\
             and pride {\Abl} {swell.\Pfv} {\Cont.\Hab} {\Cop.\Prs}\\
        \glt `And he is always swelled with pride.'
\end{exe}

\paragraph{Fossilised uses of \p{liye}} The postposition \p{liye} `because' by itself is very uncommon in modern Hindi, but its more common derivatives \p{isliye} `for this reason' and \p{kisliye} `why?' are still labelled \psst{Explanation} or \psst{Purpose}.

\hierCdef{Purpose}

\psst{Purpose} is the motivation behind an action performed (or intended to be performed) by an animate entity. \p{ke\_liye} is the prototypical adposition that takes this label. The intended outcome of an action is a \psst{Purpose}.
\begin{exe}
    \ex \gll vo bhāṣaṇ dene \p{ke\_liye} uṭhī.\\
             {\Third\Sg} {speech} {give.\Inf.\Obl} {for} {rise.\Pfv}.\\
        \glt `She rose to deliver a speech.'
\end{exe}

Intended use of something.
\begin{exe}
    \ex \label{ex:drinking}\begin{xlist}
        \ex \gll pīne \p{ke\_liye} kyā cāhiye?\\
                 {drink.\Inf.\Obl} {for} {what} {want}\\
            \glt `What do you want to drink?'
        \ex pīne \p{ko} kyā cāhiye?
    \end{xlist}
\end{exe}

Something on which an action is contingent.
\begin{exe}
    \ex \begin{xlist}
        \ex \gll sone \p{ke\_liye} koī jagah hai?\\
                 {sleep.\Inf.\Obl} {for} {any} {place} {\Cop.\Prs}\\
            \glt `Is there any space to sleep?'
        \ex film dekhne \p{ke\_liye} paise nahīṁ hai.
    \end{xlist}
\end{exe}

\paragraph{Alternations between \p{kā} and \p{ke\_liye}.} There are many cases where \p{ke\_liye} and \p{kā} are interchangeable. In such cases, it is important to check if syntactic differences arise by the exchange: \p{kā} can form a genitive PP that is a constituent of an NP, but \p{ke\_liye} obligatorily marks an adjunct. Compare:

\begin{exe}
    \ex \gll pīne \p{kā} pāni (\psst{Characteristic}; cf. \cref{ex:drinking})\\
             {drink.\Inf.\Obl} {\Gen} water\\
        \glt `drinking water'
    \ex \begin{xlist}
    \ex \gll [āne \p{ke\_liye}] [samay niścit hai]. (\psst{Purpose})\\
              {come.\Inf.\Obl} {for} {time} fixed {\Cop.\Prs}\\
        \glt `For arriving, the time is fixed.'
    \ex \gll [āne \p{kā} samay] niścit hai. (\psst{Gestalt})\\
              {come.\Inf.\Obl} {\Gen} {time} fixed {\Cop.\Prs}\\
        \glt `The arrival time is fixed.'
    \end{xlist}
\end{exe}
Or analysed using \cite{psg}'s phrase-structure grammar:

\begin{figure}[ht]
    \centering
    \begin{subfigure}{.5\textwidth}
      \centering
        \Tree [.VP [.NP [.VP \qroof{āne}.VP [.P ke ] ] [.P liye ] ] [.VP [.NP samay ] \qroof{niścit hai}.VPPred ] ]
    \end{subfigure}%
    \begin{subfigure}{.5\textwidth}
      \centering
        \Tree [.VP [.NP [.VP \qroof{āne}.VP [.P kā ] ] [.N samay ] ] \qroof{niścit hai}.VPPred ]
    \end{subfigure}
\end{figure}

\newpage
\hierAdef{Participant}

\begin{table}[ht]
    \centering
    \begin{tabular}{ll}
        \toprule
        \textbf{Verb type} & \textbf{Agent}\\
        \midrule
        Intransitive & \Nom\\
        Transitive & \Nom, \p{ne}\\
        Experiencer & \p{ko}\\
        {[Passive]} & \p{se}\\
        {[Modal]} & \p{ko}\\
        {[Nominalised]} & \p{kā}\\
        \bottomrule
    \end{tabular}
    \caption{The various markers used for the proto-Agent argument to a verb in Hindi. The verb ``classes'' in brackets are grammatical alternants of any of the verbs.}
\end{table}

\subsection{Case marker: \texorpdfstring{\p{ne}}{ne}}

As Hindi is a split-ergative language, showing both nominative--accusative and ergative--absolutive alignment, there are two primary ways to mark a canonical subject: the ergative marker \p{ne} (when the verb is in perfective aspect) or the unmarked nominative (in all other instances).

\subsubsection{Ergative}\label{sec:ergative}

\psst{Causer} is an inanimate instigator or force. Only ergative case marker \p{ne} really applies this supersense, since the kinds of entities that act as \psst{Causer}s are generally not subject to obligation, necessity, or any other modal framings that cause differential subject marking in Hindi.

\begin{exe}
    \ex \gll āg \p{ne} ghar ko naṣṭ kiyā. (\psst{Causer})\\
             fire {\Erg} house {\Acc} destroyed {do.\Pfv}\\
        \glt `The fire destroyed the home.'
\end{exe}

\psst{Agent} is the animate (or construed as such) performer of an action. The \psst{Agent} argument to a verb can be expressed with a variety of case markers depending on how the scene is to be framed.

\begin{exe}
    \ex \gll us\p{ne} kapṛe dhoye.\\
             {\Third\Sg.\Erg} {clothes.\Pl} wash.\Pfv\\
        \glt `(S)he washed clothes.'
    \ex \begin{xlist}
        \ex \gll maiṁ\p{ne} usko bahut mārā.\\
                 {\First\Sg.\Erg} {\Third\Sg.\Acc} much hit.\Pfv\\
            \glt `I hit him a lot.'
        \ex maiṁ\p{ne} usko bahut thappaṛ māre.
        \end{xlist}
\end{exe}

Verbs involving producing or creation of something (\textit{banānā} `to make'), communication (\textit{batānā} `to tell', \textit{kahnā} `to say'), and the giving of a possession (\textit{denā} `to give') take the role \rf{Originator}{Agent} for their ergative argument.

Verbs that involve a volitional experience (\textit{dekhnā} `to see', \textit{mahsūs karnā} `to feel') take the ergative. Note that these often have dative equivalent that take \rf{Experiencer}{Recipient} as their proto-Agents, e.g.~\textit{dikhāī denā} `to see'.

Verbs in which the ergative subject ends up with possession of an item (\textit{lenā} `to take', \textit{xarīdnā} `to buy') take this role.

\begin{exe}
    \ex \rf{Originator}{Agent}:\begin{xlist}
    \ex \gll maiṁ\p{ne} patra likhā.\\
             {\First\Sg.\Erg} {letter} {write.\Pfv}\\
        \glt `I wrote a letter.'
    \ex kis\p{ne} sansār ko\textsubscript{\psst{Theme}} banāyā?
    \ex maiṁ\p{ne} āpko\textsubscript{\psst{Recipient}} tohfā diyā.
    \end{xlist}
    \ex \rf{Experiencer}{Agent}:\begin{xlist}
    \ex \gll ham\p{ne} khelte hue baccoṁ ko dekhā.\\
             {\First\Pl.\Erg} {play.\Hab} {\Cop.\Pfv} {children.\Obl} {\Acc} {see.\Pfv}\\
        \glt `We saw children playing.'
    \ex maiṁ\p{ne} dhyān se\textsubscript{\psst{Manner}} sunā.
    \end{xlist}
    \ex \rf{Recipient}{Agent}:\begin{xlist}
    \ex \gll Rām \p{ne} mujhse\textsubscript{\rf{Originator}{Source}} kitāb le lī.\\
             {Ram} {\Erg} {\First\Sg.\Abl} {book} {take} {take.\Pfv}\\
        \glt `Ram took the book from me.'
    \end{xlist}
    \ex \rf{SocialRel}{Agent}:\begin{xlist}
    \ex maiṁ\p{ne} tumse śādī karnī hai.
    \end{xlist}
\end{exe}

\paragraph{Bodily emission verbs.} There is a set of `bodily emission' verbs \citep{dehoop2005dam}, such as \textit{chīṁknā} `to sneeze', \textit{khāṁsnā} `to cough', \textit{mūtnā} `to urinate', that can optionally take the ergative marker (sometimes with light verb constructions) for their subject. The presence of the marker indicates greater agency, so we treat it as \psst{Agent} (the lack of the marker would make it a \psst{Theme}). Note that this alternation is not permissible for every speaker.\footnote{This specific example doesn't work for Aryaman, but e.g.~\w{cīkhnā} `to yell' does.}

\begin{exe}
    \ex \begin{xlist}
            \ex \gll us\p{ne}\textsubscript{\psst{Agent}} chīṁkā\\
                 {\Third\Sg.\Obl.\Erg} {sneeze.\Pfv}\\
            \glt `He sneezed [on purpose].'
            \ex \gll vo\textsubscript{\psst{Theme}} chīṁkā\\
                 {\Third\Sg} {sneeze.\Pfv}\\
            \glt `He sneezed [involuntarily].'
        \end{xlist}
\end{exe}

\paragraph{Non-agentive ergatives.} There is also a set of verbs that obligatorily take the ergative (as well as the usual modal alternations) even when forming inherently non-agentive compound verbs. Noun-verb concatentions with \w{khānā} `to eat' behave this way, apparently with a figurative extension of `eat' to `receive' or `bear'. 

\begin{exe}
    \ex \gll maiṁ\p{ne} us\punbold{se}\textsubscript{\rf{Originator}{Source}} mār khāyī.\\
             {\First\Sg.\Erg} {\Third\Sg.\Obl.\Abl} beating {eat.\Pfv}\\
        \glt I took a beating from him.
\end{exe}

Since we annotate source domain of metaphors, the subject should be annotated \rf{Recipient}{Agent} here.\footnote{This is an open issue, \ghi{1}.}

\begin{discussion}
In the differentially-marked subjects for obligation, necessity, and ability, the \psst{Agent}s do not have volition, so that scene role for them is uncertain. This is part of the broader problem of SNACS's treatment of force dynamics cross-lingually, and will not be easily resolved with the current hierarchy. 
\end{discussion}

\subsection{Case marker: \texorpdfstring{\p{ko}}{ko}}

Like in most Indo-Aryan languages, \p{ko} is a dative--accusative marker. Both senses seem to constitute a single entry in the lexicon; the difference between a dative \p{ko} and an accusative \p{ko} is not readily known to a non-linguistically-informed native speaker.

\paragraph{Syntactic tests for ascertaining function.} The dative \p{ko} is obligatory while the accusative \p{ko} marks animacy, definiteness, and/or salience. Thus, one can use an indefinite (e.g.~a plural) and/or inanimate substitution  to test if the \p{ko} can be dropped; if it can be, then it is an accusative.

\begin{exe}
    \ex Accusative:\begin{xlist}
    \ex {[mez \p{ko}]}\textsubscript{\psst{Theme}} sāf karo.
    \ex {[bīs mez]}\textsubscript{\psst{Theme}} sāf karo.
    \end{xlist}
    \ex Dative (dropping \p{ko} changes the role):\begin{xlist}
    \ex us\p{ko}\textsubscript{\psst{Recipient}} dikhāo.
    \ex vah\textsubscript{\rf{Stimulus}{Theme}} dikhāo.
    \end{xlist}
\end{exe}
See also \citet[72--76]{psg}.

\subsubsection{Accusative (and \texorpdfstring{\p{kā}}{ka}, \texorpdfstring{\p{par}}{par})}\label{sec:accusative}

The various accusative markers are all annotated \psst{Theme}. A \psst{Theme} undergoes an action, nonagentive motion, a change of state, or transfer. It is a broad category, best signified by the differentially marked (generally on animate or specific objects) accusative \p{ko}. Some compound verbs favour \p{kā} or \p{par} as their object markers.

The pronouns have special accusative forms suffixed with \w{-e(ṁ)} (\textit{mujhe}, \textit{tujhe}, \textit{hameṁ}, \textit{use}, etc.), which are all treated the same as \p{ko}.

\begin{exe}
    \ex \begin{xlist}
        \ex \gll mez \p{ko} sāf karo.\\
                 table {\Acc} clean {do.\Imp}\\
            \glt `Clean the table.'
        \ex \gll mez \p*{kī}{kā} safāī karo.\\
                 table {\Gen} cleaning {do.\Imp}\\
            \glt `Do the cleaning of the table.'
        \end{xlist}
    \ex \gll usne bacce \p{ko} sulāyā.\\
             {\Third\Sg.\Erg} {child.\Obl} {\Acc} {sleep.\Caus.\Pfv}\\
        \glt `(S)he made the child sleep.'
    \ex arjun ne mahābhārat meṁ karṇ \p{ko} parājit kiyā.
    \ex \gll usne kitāb \p{ko} becā.\\
             {\Third\Sg.\Erg} book {\Acc} {sell.\Pfv}\\
        \glt `(S)he sold the book.'
    \ex maiṁne \p*{use}{ko} ḍākghar bhejā.
    \ex us\p*{kī}{kā} piṭāī
\end{exe}
Some verbs use \p{par}.
\begin{exe}
    \ex \gll hamne tum \p{par} hamlā kiyā.\\
             {\First\Pl.\Erg} {\Second\Pl} {on} attack {do.\Pfv}\\
        \glt `We attacked you.'
    \ex maiṁne is deś \p{par} rāj kiyā.
\end{exe}

Other examples of \p{ko} marking verbal arguments are below. \rf{Stimulus}{Theme} marks the source of a volitional experience, such as \textit{dekhnā} `to see', \textit{sunnā} `to hear'. Some verbs (\w{samajhnā} `to understand', \w{mānnā} `to accept', etc.) license a \rf{Topic}{Theme} for their objects (\ghi{3}). This includes the adjective--verb compound use of \textit{samajhnā}.

\begin{exe}
    \ex \rf{Stimulus}{Theme}:\begin{xlist}
    \ex \gll maiṁ baccoṁ \p{ko} dekh rahā thā.\\
             {\First\Sg} {child.\Pl.\Obl} {\Acc} see {\Cont} {\Cop.\Pst}\\
        \glt `I was watching a movie.'
    \end{xlist}
    \ex \rf{Topic}{Theme}:\begin{xlist}\ex\gll jīvan \p{ko} samajhnā muśkil hai.\\
             life {\Acc} {understand.\Inf} difficult {be.\Prs}\\
        \glt `Understanding life is difficult.'
    \ex \gll kyā tum mujh\p*{e}{ko} ullū samajhte ho?\\
            {what} {\Second\Pl} {\First\Sg.\Acc} owl {understand.\Hab} {\Cop.\Prs}\\
        \glt 'Do you think I'm stupid?'
    \end{xlist}
    \ex \rf{Possession}{Theme}:\begin{xlist}
    \ex is gande kele \p{ko} nahīṁ xarīdūṁgā!
    \end{xlist}
\end{exe}

\paragraph{Spray--load alternation.} Like English (and many other languages), Hindi exhibits a \textit{spray--load alternation} that allows \p{ko} to take the construal \rf{Goal}{Theme}.\footnote{This \href{https://twitter.com/aryaman2020/status/1337555961750450176}{informal Twitter poll} ($n = 45$) finds that $29\%$ of respondents think the alternation means different things, with the form marking the container with the accusative implying `filling to the top'. The form marking the liquid with the accusative is greatly preferred ($62\%$).}

\begin{exe}
    \ex \begin{xlist}
    \ex \label{ex:sprayloadins}\gll gilās \p{ko} pāni \punbold{se}\textsubscript{\rf{Theme}{Instrument}} bharo. (\rf{Goal}{Theme})\\
             glass {\Acc} water {\Ins} {fill.\Imp}\\
        \glt `Fill the glass with water.'
    \ex \gll gilās \punbold{meṁ}\textsubscript{\rf{Goal}{Locus}} pāni bharo.\\
             glass {\Acc} water {\Ins} {fill.\Imp}\\
        \glt `Fill the water in the glass.'
    \ex \gll pānī \p{ko} gilās \punbold{meṁ}\textsubscript{\rf{Goal}{Locus}} bharo. (\psst{Theme})\\
             water {\Acc} glass {\Loc} {fill.\Imp}\\
        \glt `Fill the water in the glass.'
    \end{xlist}
\end{exe}

\begin{discussion}
\psst{Theme}-type markers are often used to mark the object of a verb (such as a causative) with force-dynamic properties.
\begin{exe}
    \ex \gll aurat ne bacce \p{ko} sulāyā.\\
             woman {\Erg} {child.\Obl} {\Acc} {sleep.\Caus.\Pfv}.\\
        \glt `The woman made the child sleep.'
\end{exe}
\psst{Theme} is perhaps not the best label for this, but since there is no special handling of force dynamics, this is the best option in the current hierarchy.
\end{discussion}

\subsubsection{Dative}\label{sec:dative}

The function for this case is \psst{Recipient}. The canonical example of the dative is an indirect object to which the direct object (\psst{Theme}) is transferred by the subject (\psst{Originator}).\footnote{In Universal Dependencies \citep{nivre2020universal}, the \psst{Recipient} is the argument to the verb that has an \texttt{iobj} relation.}

\begin{exe}
    \ex \gll maiṁne apne dost \p{ko} kitāb dī.\\
             {\First\Sg.\Erg} {\Refl.\Gen} {friend} {\Dat} book {give.\Pfv}\\
        \glt `I gave my friend the book.'
    \ex \gll mujh\p*{e}{ko} ek bāt batāo.\\
             {\First\Sg.\Dat} one talk {tell.\Imp}\\
        \glt `Tell me one thing.'
    \ex \gll Suṣmitā \p{ko} kitne pāṭh paṛhāoge?\\
             {Sushmita} {\Dat} {how.many} lessons {teach.\Fut}\\
        \glt `How many lessons will you teach to Sushmita?'
    \ex tum\p{ko} ek cīz dikhānā cāhtā hūṁ. (\rf{Experiencer}{Recipient})
\end{exe}

\paragraph{Dative subject.} The dative subject (sometimes narrowly called the experiencer subject) is a common construction with some verbs in Hindi. Besides \psst{Experiencer}s, it also marks some idiosyncratic verbs.

\begin{exe}
    \ex \rf{Experiencer}{Recipient}: \begin{xlist}
    \ex \gll Sunītā \p{ko} buxār hai.\\
             Sunita {\Dat} fever {\Cop.\Prs}\\
        \glt `Sunita has a fever.'
    \ex \gll mujh\p{ko} Hindī nahīṁ ātī.\\
             {\First\Sg.\Dat} {Hindi} {\Neg} {come.\Hab}\\
        \glt `I don't know Hindi.'
    \ex rājā \p{ko} duḥkh huā.
    \ex mujh\p{ko} tum pasand ho.
    \ex ek ām \p{ko} dūsre ām se\textsubscript{\rf{Stimulus}{Ancillary}} pyār huā.
    \ex us\p{ko} acānak se\textsubscript{\psst{Manner}} āvāz sunāī dī.
    \ex mujh\p*{e}{ko} dūsrī kitāb cāhiye.
    \ex Amerikā \p{ko} koī aitrāz nahīṁ hai.
    \end{xlist}
    \ex \rf{Beneficiary}{Recipient}:\begin{xlist}
    \ex \gll mujh\p*{e}{ko} koī fāydā nahīṁ huā.\\
             {\First\Sg.\Dat} any benefit {\Neg} {\Cop.\Pfv}\\
        \glt `I got no benefit.'
    \ex kampanī \p{ko} munāfā hogā.
    \ex māṁg \p{ko} Kāṁgres kā samarthan hai.
    \end{xlist}
    \ex \rf{Gestalt}{Recipient}:\begin{xlist}
    \ex \gll ām ādmī \p{ko} haq hai.\\
             common man {\Dat} right {\Cop.\Prs}.\\
        \glt `The common man has the right.'
    \ex mujh\p{ko} bahut kām hai.
    \end{xlist}
    \ex \rf{SocialRel}{Recipient}:\begin{xlist}
    \ex \gll Rām \p{ko} do beṭiyāṁ huī.\\
             Ram {\Dat} two {daughter.\Pl} {be.\Pfv}.\\
        \glt `Two daughters were born to Ram.'
    \end{xlist}
\end{exe}

\paragraph{Modal subject.} In conjunction with some modal light verbs, the dative case marker \p{ko} marks the \psst{Agent}. These, however, have force dynamic issues.

\begin{exe}
    \ex \gll us\p{ko} pānī pīnā cāhiye. (\rf{Agent}{Recipient}) \\
             {\Third\Sg.\Obl.\Dat} {water} {drink.\Inf} {should} \\
        \glt `He should drink water.'
    \ex \gll rām \p{ko} kitāb band karnī paṛī. (\rf{Agent}{Recipient}) \\
             Ram {\Dat} book close {do.\Inf} {be.obliged.\Pfv} \\
        \glt `Ram had to close the book.'
\end{exe}

\subsection{\texorpdfstring{\psst{Topic}: \p{ke\_bāre\_meṁ}, etc.}{Topic: ke\_bare\_mem, etc.}}\label{sec:topic}

\psst{Topic} primarily refers to information content (especially in cognition event) or communication. The prototypical adposition for this is \p{ke\_bāre\_meṁ}, but the locative markers \p{meṁ}, \p{par} and genitive \p{kā} mark \psst{Topic}s as well.

\begin{exe}
    \ex \gll \p*{tumhāre\_bāre\_meṁ}{ke\_bāre\_meṁ} bāt kar rahe the.\\
             {\Second\Pl.about} talk do {\Cont} {\Cop.\Pst}\\
        \glt `We were talking about you.'
    \ex \gll us\p*{kī}{kā} tasvīr dikhāo.\\
             {\Third\Sg.\Gen} {picture} {show.\Imp}\\
        \glt `Show the picture of him.'
    \ex \gll te\p*{rā}{kā} kyā hogā?\\
             {\Second\Sg.\Gen} {what} {\Cop.\Fut}?\\
        \glt `What will become of you?'
    \ex is bāt \p{par} carcā huī.
\end{exe}

\subsection{Case marker: \texorpdfstring{\p{se}}{se}}

\subsubsection{Instrumental (and \texorpdfstring{\p{ke\_zariye}}{ke\_zariye} etc.)}\label{sec:instrumental}

The instrumental case (\Ins) of \p{se} takes the function \psst{Instrument}.

\begin{exe}
    \ex \gll maiṁne cāqū \p{se} sabzī ko kāṭā.\\
             {\First\Sg.\Erg} {knife} {\Ins} vegetable {\Acc} {cut.\Pfv}\\
        \glt `I cut the vegetables with a knife.'
    \ex \gll gāṛī \p{se} ghar jāūṁgā.\\
              {car} {\Ins} home {go.\Fut}\\
        \glt `I'll go home by car.'
    \ex tohfe ko ḍāk \p{se} bhejo.
    \ex us rāste \p{se} jāo. (\rf{Path}{Instrument})
    \ex gilās ko pāni \p{se} bharo. (\rf{Theme}{Instrument}, from \cref{ex:sprayloadins})
\end{exe}

The postpositions \p{ke\_zariye} `via, through' and \p{ke\_mādhyam\_se} `by means of' (in Sanskritised Hindi) also mark \psst{Instrument}s.

\begin{exe}
    \ex Gūgal \p{ke\_zariye} khoj lo.
    \ex Hindī bhāṣā \p{ke\_mādhyam\_se} ham logoṁ tak pahuṁc sakte haiṁ.
\end{exe}

\paragraph{Animate instruments.} Indirect causative verbs in Hindi (e.g.~\textit{khulvānā} `to make X open Y') can take an animate instrument which exhibits \psst{Agent}-like properties \citep{se-causative}. Currently we annotate these as their predicate-licensed scene role construed as \psst{Instrument}.

\begin{exe}
    \ex \gll maiṁne bāī \p{se} bacce ko sulvāyā. (\rf{Agent}{Instrument})\\
             {\First\Sg.\Erg} {maid} {\Ins} {child.\Obl} {\Acc} {sleep.\Caus2.\Pfv}\\
        \glt `I made the maid put the child to sleep.'
    \ex tumne mujh\p{se} khānā banvāyā (\rf{Originator}{Instrument})
\end{exe}

\begin{discussion}
One possible change to this is to create a new function for animate instruments: \textsc{Aider}. Animate instruments can control adverbial phrases while inanimate instruments cannot, animate instruments can control instruments of their own, and a similar distinction already exists between inanimate \psst{Causer} and animate \psst{Agent} in the hierarchy \citep{bhatia2016,begum-sharma-2010-preliminary}, thus it seems strange to say these are still morphosyntactic \psst{Instrument}s.

An alternative is to treat this as an \psst{Agent} and make a new supersense for the initator of the action (which is a volition entity but not an actor itself). This approach is taken by Bill Croft.\footnote{Personal communication.}
\end{discussion}

\subsubsection{Ablative}\label{sec:ablative}

The ablative sense of \p{se} (\Abl) takes the function \psst{Source}. (For the literal meaning of motion away, see that section.)

Some of the literal ablative uses to mark verbal arguments get the scene role \psst{Theme}; refer to the English guidelines \citep{en} for more on this.

\begin{exe}
    \ex \gll adhyāpak ne laṛkoṁ \punbold{ko}\textsubscript{\psst{Theme}} laṛkiyoṁ \p{se} alag kiyā. (\rf{Theme}{Source})\\
             teacher {\Erg} {boy.\Pl.\Obl} {\Acc} {girl.\Pl.\Obl} {\Abl} separate {do.\Pfv}\\
        \glt `The teacher separated the boys from the girls.'
\end{exe}

Here are some of the more grammaticalised uses of ablative \p{se} to mark verbal arguments, classified as such based on typological considerations given in \citep{tafseer}. The counter-intuitive \rf{Recipient}{Source} frames the \psst{Recipient} of a `request'-type verb as the \psst{Source} of a response. 

\begin{exe}
    \ex \rf{Stimulus}{Source}: \begin{xlist}
    \ex \gll tum\p{se} ḍar lagtā hai.\\
             {\Second\Pl.\Abl} {fear} {feel.\Hab} {\Cop.\Prs}\\
        \glt `I feel scared of you.'
    \ex tumhāre bartāv \p{se} maiṁ ġussā hūṁ.
    \ex maiṁne āp\p{se} ummīd rakhī.
    \ex pyār huā, iqrār huā hai, pyār \p{se} phir kyoṁ ḍartā hai dil?
    \end{xlist}
    \ex \rf{Recipient}{Source}: \begin{xlist}
    \ex \gll maiṁ āp\p{se} bhīk māṁgtā hūṁ.\\
             {\First\Sg} {\Second\Pl.\Abl} {alms} {ask.for.\Hab} {\Cop.\Prs}\\
        \glt `I beg of you.'
    \ex usne mujh\p{se} praśn pūchā.
    \end{xlist}
    \ex \rf{Originator}{Source}: \begin{xlist}
    \ex \gll kyā tumheṁ us\p{se} kuch milā?\\
             {what} {\Second\Pl.\Dat} {\Third\Sg.\Abl} anything {receive.\Pfv}\\
        \glt `Did you get anything from them?'
    \ex dost \p{se} patā calā.\footnote{But note that an inanimate provider of information (like a book) is just \psst{Source}. See issue \ghi{20}.}
    \end{xlist}
    \ex \rf{Causer}{Source}: \begin{xlist}
    \ex \gll vah zukām \p{se} pīṛit hai.\\
             {\Third\Sg} {cold} {\Abl} suffering {\Cop.\Prs}\\
        \glt `He is suffering from the cold.'
    \end{xlist}
\end{exe}

\subsubsection{Comitative (and \texorpdfstring{\p{ke\_sāth}}{ke\_sath}, \texorpdfstring{\p{ke\_binā}}{ke\_bina})}\label{sec:comitative}

The comitative sense of \p{se} (\Com) takes the function \psst{Ancillary} and usually a non-matching scene role falling under \psst{Participant}. Verbs, pertaining to social relations (\textit{śādī karnā}) and emotional stimuli (\textit{pyār/nafrat karnā}), as well as more literal verbs involving association or joining, license the comitative \p{se} for the secondary participant:

\begin{exe}
    \ex \gll vah tum\p{se} pyār kartī hai. (\rf{Stimulus}{Ancillary})\\
             {\Third\Sg} {\Second\Pl.\Com} love {do.\Hab} {\Cop.\Prs}\\
        \glt `She loves you.'
    \ex \gll ham un\p{se} ṭakrāye. (\rf{Theme}{Ancillary})\\
             {\First\Pl} {\Third\Pl.\Com} {collide.\Pfv}\\
        \glt `We collided with them.'
    \ex kyā tum mujh\p{se} śādī karogī beġam? (\rf{SocialRel}{Ancillary})
\end{exe}

Just like English has a distinction between accompaniers (\textit{together with}) and secondary participants in a scene (\textit{with}), Hindi distinguishes \p{ke\_sāth} and the comitative sense of \p{se}. For example:

\begin{exe}
    \ex \begin{xlist}
    \ex \gll maiṁ tum\p{se} laṛūṅgā. (\rf{Agent}{Ancillary})\\
             {\First\Sg} {\Second\Pl.\Com} {fight.\Fut}\\
        \glt `I will fight you.'
    \ex \gll maiṁ \p*{tumhāre\_sāth}{ke\_sāth} laṛūṅgā. (\psst{Ancillary})\\
             {\First\Sg} {\Second\Pl.with} {fight.\Fut}\\
        \glt `I will fight with you.'
    \end{xlist}
\end{exe}

\paragraph{Behaviour.} The postposition \p{ke\_sāth} can be used with certain predicates to indicate the target of behaviour. Following the main guidelines, we label this \rf{Beneficiary}{Ancillary} (see \ghi{9}).

\begin{exe}
    \ex \gll tumne \p*{mere\_sāth}{ke\_sāth} burā bartāv kiyā.\\
             {\Second\Pl.\Erg} {\First\Sg.with} bad behaviour do.\Pfv\\
        \glt `You behaved poorly with me.'
\end{exe}

\begin{discussion}
There are some cases where \p{ke\_sāth} and \p{se} are interchangeable, probably given the relatively recent grammaticalisation of \p{ke\_sāth} as a comitative. One instance is the second argument to \textit{khelnā} `to play', which can take either case if the argument is inanimate:
\begin{exe}
    \ex \begin{xlist}
    \ex maiṁ guṛiyā \p{se} khel rahā hūṁ.
    \ex \gll maiṁ guṛiyā \p{ke\_sāth} khel rahā hūṁ.\\
             {\First\Sg} {doll} {with} play {\Cont} {\Cop.\Prs}\\
        \glt `I am playing with a doll.'
    \end{xlist}
    \ex \begin{xlist}
    \ex *maiṁ dost \p{se} khel rahā hūṁ.
    \ex \gll maiṁ dost \p{ke\_sāth} khel rahā hūṁ.\\
             {\First\Sg} {friend} {with} play {\Cont} {\Cop.\Prs}\\
        \glt `I am playing with a friend.'
    \end{xlist}
\end{exe}
Only \p{ke\_sāth} is grammatical for an animate.\footnote{Based on native speaker judgement. See \href{https://twitter.com/aryaman2020/status/1356987346164584448}{this Twitter poll}: $n = 48$, \p{ke\_sāth} is preferred by $92\%$ for an animate, but only $52\%$ for an inanimate.} For an inanimate participant, arguably the scene role \psst{Agent} is impossible (and there is no personification at play), so we label \p{ke\_sath} as \psst{Ancillary} and \p{se} as \psst{Instrument}.
\end{discussion}

\subsubsection{Passive subject (and \texorpdfstring{\p{dvārā}}{dvara})}\label{sec:passivesubject}

A verb in a passive construction (with the light verb \textit{jānā} ``to go'') marks the \psst{Agent} (with appropriate predicate-licensed scene role) with the instrumental case marker \p{se}. This can also be a debilitative construction when negated.

The postposition \p{dvārā} also marks a passive subject in some dialects and literary Hindi.

\begin{exe}
    \ex \gll bacce \p{se} śīśā ṭūṭ gayā. (\psst{Agent})\\
             {child.\Obl} {\Ins} mirror break {go.\Pfv}\\
        \glt `The mirror was broken by the child.'
    \ex \gll Rām \p{dvārā} likhit (\psst{Agent})\\
             Ram by written\\
        \glt `written by Ram'
    \ex mujh\p{se} nā \choices{kiyā jāyegā\\ho pāyegā}. (\psst{Agent})
\end{exe}

\begin{discussion}
Given that the passive is thoroughly grammaticalised into Hindi, it is not apparent which canonical case it belongs under. The most likely candidates are the ablative (\psst{Source}) or the instrumental (\psst{Instrument}); compare the Sanskrit instrumental and genitive being used in a similar way with participles historically. Given these facts, we elected to make \psst{Agent} a valid function for \p{se}.
\end{discussion}

\subsection{Case marker: \texorpdfstring{\p{kā}}{ka}}\label{sec:genitive}

The main use of \p{kā} to mark a \psst{Participant} is in nominalisations of verb phrases, in which it marks arguments to the verb.

\begin{exe}
    \ex \gll Sacin Tendulkar \p{kā} 200 ran \punbold{kā}\textsubscript{\psst{Identity}} {rikarḍ} (\rf{Agent}{Gestalt})\\
             Sachin Tendulkar {\Gen} 200 run {\Gen} {record}\\
        \glt `Sachin Tendulkar's 200 run record'
    \ex \p*{tumhārā}{kā} us\punbold{ko}\textsubscript{\psst{Theme}} mārnā (\rf{Agent}{Gestalt})
    \ex \gll \p*{merī}{kā} {samajh} {meṁ} {āyā} (\rf{Experiencer}{Gestalt})\\
             {1\Sg.\Gen} {understanding} {\Loc} {come.\Pfv}\\
        \glt `I understood.'
    \ex \gll \p*{tumhārā}{kā} yah likhnā ṭhīk nahīṁ thā. (\rf{Originator}{Gestalt}) \\
             {\Second\Pl.\Gen} {\Third\Sg} {write.\Inf} {proper} {\Neg} {\Cop.\Pst}\\
        \glt `You writing this was not okay.'
    \ex \gll Sītā \p*{kī}{kā} hāmī (\rf{Originator}{Gestalt})\\
             Sita {\Gen} assent\\
        \glt `Sita's assent.'
\end{exe}



\subsection{Postposition: \texorpdfstring{\p{ke\_liye}}{ke\_liye}}\label{sec:benefactive}

\p{ke\_liye} when marking something animate it indicates a \psst{Beneficiary}.

\begin{exe}
    \ex \gll maiṁne \p*{tumhāre\_liye}{ke\_liye} kiyā.\\
             {\First\Sg.\Erg} {\Second\Pl.for} {do.\Pfv}\\
        \glt `I did it for you.'
\end{exe}

It is similar to English \textit{for}; it can also indicate purposes (see \psst{Purpose}), sufficiency/excess comparisons (see \psst{ComparisonRef}), and costs.

\begin{exe}
    \ex \gll \p*{iske\_liye}{ke\_liye} do rupaye lageṅge. (\psst{Cost}) \\
             {\Third\Sg.for} two {rupee.\Pl} {apply.\Fut}\\
        \glt `This will cost two rupees.'
\end{exe}

It is also used in a manner similar to the English \textit{from the perspective of}, where it licenses \rf{Experiencer}{Beneficiary}.

\begin{exe}
    \ex \p*{mere\_liye}{ke\_liye} bahut āsān kām hai. (\rf{Experiencer}{Beneficiary})
\end{exe}

\subsection{Postposition: \texorpdfstring{\p{ke\_xilāf}}{ke\_xilaf}, \texorpdfstring{\p{ke\_viruddh}}{ke\_viruddh}}\label{sec:against}

The postpositions \p{ke\_xilāf} and \p{ke\_viruddh} (in Sanskritised Hindi) indicates a maleficiary, which is classified as a type of \psst{Beneficiary} in SNACS. They are similar to English \textit{against}.

\begin{exe}
    \ex paramparā \p{ke\_viruddh}
    \ex \gll kis deś \p{ke\_xilāf} yuddh hogi? (\rf{Agent}{Beneficiary})\\
             which country against war {be.\Fut}\\
        \glt `Against which country will the war be fought?'
    \ex faisle \p{ke\_xilāf} honā (\rf{Characteristic}{Beneficiary})
\end{exe}

\subsection{Postposition: \texorpdfstring{\p{ke\_binā}}{ke\_bina}}\label{sec:without}

When marking an animate NP and as an adjunct to a verb, \p{ke\_binā} `without' is labelled \psst{Ancillary} (i.e. it is the negation of \p{ke\_sāth}. It also has a unique syntactic variant that is circumpositional: \p*{binā\_ \_ke}{ke\_binā}, which is generally utilised with inanimates. See \ghi{23} for some corpus examples.

\begin{exe}
    \ex \gll kyā tum \p*{mere\_binā}{ke\_binā} dukān jā sakte ho?\\
             what {\Second\Pl} {\First\Sg.without} store go {be.able.\Hab} {\Cop.\Prs}\\
        \glt `Can you go to the store without me?'
\end{exe}

For inanimates, it is labelled \rf{Possession}{Ancillary}. To check for this, you may test if the opposite meaning with the conjunctive verb \textit{lekar} is valid (e.g.~\textit{āp vīzā lekar...}).

\begin{exe}
    \ex \gll āp \p*{binā\_}{ke\_binā} vīzā \p*{\_ke}{ke\_binā} nahīṁ jā sakte.\\
             {\Second\Pl} {without} vise {\Gen} {\Neg} go {be.able.\Hab}\\
        \glt `You cannot go without a visa.'
\end{exe}

When the object is an action (whether nominal or verbal), then \p{ke\_binā} could be labelled either \psst{Circumstance} or \psst{Manner}. If it can answer a \textit{kaise?} question then it is the latter.

\begin{exe}
    \ex \gll tum \p{binā} batāye cal gaye? (\psst{Circumstance})\\
             {\Second\Pl} {without} {tell.\Pfv} walk {go.\Pfv}\\
        \glt `You left without telling?'
    \ex \p*{binā\_}{ke\_binā} kisī kī madad \p*{\_ke}{ke\_binā} (\psst{Manner})
    \ex \p*{binā\_}{ke\_binā} ghaṭnā \p*{\_ke}{ke\_binā} (\psst{Manner})
\end{exe}

\hierAdef{Configuration}

\hierBdef{Identity}

\p{kā} (\Gen) can be used to categorise or equate, thus being labelled \psst{Identity}.

\begin{exe}
    \ex \gll use maut \p*{kī}{kā} sazā milegī.\\
             {\Third\Sg.\Dat} {death} {\Gen} {punishment} {receive.\Fut}\\
        \glt `(S)he will be sentenced to death.'
    \ex cār sāl \p*{kī}{kā} umr
    \ex \gll ānand ṭelīfon ŏpreṭar \p{kā} kām kartā hai.\\
             Anand {telephone} {operator} {\Gen} {work} {do.\Hab} {\Cop.\Prs}\\
        \glt `Anand works as a telephone operator.'
\end{exe}

\p{ke\_rūp\_meṁ}, analogous to the English construction \textit{as}, takes an object (core argument) of a predicate and categorises it with a label that bears the postposition.

\begin{exe}
    \ex \gll {Yuvaraj} {acchī} {krikeṭar} \p{ke\_rūp\_meṁ} {pari-pakv\u a} {ho cuke} hai.\\
    {Yuvaraj} {good} {cricketer} {as} {mature} {complete.\Pfv} {\Cop.\Prs}\\
    \glt `Yuvaraj has matured as a good cricketer.'
    
    \ex 14 sitambar kā din 'hindi-divas' \p{ke\_rūp\_meṁ} manāyā jātā hai.\\
\end{exe}

\hierBdef{Species}

\psst{Species} is rare in Hindi. The main instance of this is when the governor of the \p{kā}-marked NP is a word like \textit{misāl} or \textit{udāhraṇ} `example'.

\begin{exe}
    \ex \gll Bhār\u{a}tīy kalā \p{kā} udāhraṇ\\
             Indian art {\Gen} example\\
        \glt `an example of Indian art'
\end{exe}

\paragraph{Confusion with \psst{Characteristic}.}
Semantically, the usual translation equivalent of English \textit{type of X} into Hindi is \textit{tarah kā X}. Note, however, that the head of this NP is opposite in Hindi: it is \textit{X} rather than \textit{type}. That construction with \p{kā} is labelled \psst{Characteristic}.

\begin{figure}[ht]
    \centering
    \begin{subfigure}{.45\textwidth}
      \centering
        \Tree [.NP [.D that ] [.N' [.N kind ] [.PP [.P of ] [.NP [.N friend ] ] ] ] ]
    \caption{A representation of English \textit{that kind of friend} in $\overline{X}$ theory.}
    \end{subfigure}\hspace{.05\textwidth}
    \begin{subfigure}{.45\textwidth}
      \centering
        \Tree [.NP [.NP [.NP [.Dem us ] [.N tarah ] ] [.P kā ] ] [.NP [.N dost ] ] ]
    \caption{A representation of Hindi \textit{us tarah kā dost} per \citet{psg}.}
    \end{subfigure}
\end{figure}

\hierBdef{Gestalt}

\psst{Gestalt} is the prototypical function of \p{kā}, and the genitive forms of pronouns (e.g.~\textit{merā} `\First\Sg.\Gen'). Note that the genitives are declined for the gender of their governor.

\begin{exe}
    \ex \gll \p*{merā}{kā} nām Rām hai.\\
             {\First\Sg.\Gen} name Ram {\Cop.\Prs}\\
        \glt `My name is Ram.'
    \ex ām \p{kā} dām
    \ex kām karne \p{kā} nayā tarīqā
    \ex \gll ve TV ke\_sāth \p*{apnā}{kā} samay bitāte haiṁ\\
    {3\Pl} TV {with} {\Refl.\Gen} {time} {spend.\Hab} {\Cop.\Prs}\\
    \glt `They spend their time with the TV.'
    \ex \gll {dūdh} \p{kī} {miṭhās} {acchī} {hai} \\
    {milk} {\Gen} {sweetness} {good} {\Cop.\Prs} \\
    \glt `The milk is sweet.' [lit. `The milk's sweetness is good.']
\end{exe}

For \psst{Gestalt}, possession is typically complex or abstract, and usually not alienable (otherwise \psst{Possessor} is used).

As a function, it is also used for nominalisations of verb phrases.

\paragraph{Possessive \p{ke\_pās}.} Like in many Indo-Aryan languages, the postposition for `near' (\p{ke\_pās}) has come to have a possessive sense. This is labelled \rf{Gestalt}{Locus} (or with a subtype scene role). It was elected not to give the function \psst{Gestalt} to this since it often implies physical on-person possession when contrasted with the genitive \p{kā}.

\begin{exe}
\ex \gll{paṛhai} {ke kāraṇ} {us\p{ke\_pās}} {samay} {nahin} {hai} (\rf{Gestalt}{Locus})  \\
{studies} {\Caus} {3SG.\Loc} {time} {\Neg} {\Cop.\Sg}\\
\glt `He has no time on account of (his) studies.
\end{exe}

\paragraph{Locative subject alternation.} The locative case marker \p{meṁ}, when applied to a subject of a verb, can indicate a \rf{Gestalt}{Locus}, the possessor of a property \citep{kachru1970}.

\begin{exe}
    \ex \begin{xlist}
    \ex laṛke \p{kā} sāhas (\psst{Gestalt})\\
        {boy.\Obl} {\Gen} courage\\
        `the courage of the boy'
    \ex laṛke \p{meṁ} sāhas hai (\rf{Gestalt}{Locus})\\
        {boy.\Obl} {\Loc} courage {\Cop.\Prs}\\
        `The boy is courageous.'
    \end{xlist}
    \ex merī\textsubscript{\rf{Agent}{Gestalt}} harkatoṁ \p{meṁ} pyār hai. (\rf{Gestalt}{Locus})
    \ex \p*{mere\_pās}{ke\_pās} māṁ hai. (\rf{SocialRel}{Locus})
\end{exe}

\hierCdef{Possessor}
The \psst{Possessor} label is again associated with genitive \p{kā}. This is only for alienable possessions of property (generally physical item, but also less tangible property like data or Bitcoins).

Like in English, this includes possessions implying but not explicitly stating previous transfer events.

\begin{exe}
\ex \gll yah kis\p{kā} paisā hai \\
{\Third\Sg} {who.\Gen} {money} {\Cop.\Prs} \\
\glt `Whose money is this?'

\ex \gll kal \p*{tumhārī}{kā} {chiṭṭhī} {āyī} {thī} \\
{yesterday} {you.\Obl.\Gen.\Sg}  {letter} {come.PERF.\Pst} {\Cop.\Pst}\\
\glt `Your letter had arrived yesterday.'

\ex \gll laṛke \p{ke\_pās} {paise} nahīṁ hai. (\rf{Possessor}{Locus}) \\
{boy.\Obl} {near} {money} {\Neg} {\Cop.\Prs} \\
\glt `The boy does not have money.'

\ex \p*{merā}{kā} ḍīṅgā ām bahut acchā hai.
\end{exe}

\hierCdef{Whole}
\psst{Whole} largely follows the English guidelines (\citealt{en}) in its definitions for Hindi, associated chiefly with the genitive \p{kā} and the locative \p{meṁ} in constructions with the copula \citep{kachru1970}.

The possessed entity is well-defined on its own, yet not alienable in the sense of being unable to exist by its own self:

\begin{exe}
\ex \gll {ādmi} {ghar} \p*{ke}{kī} {chat} {par} {baiṭhā} {hai} (\psst{Whole}) \\
{man} {house} {\Gen} {roof} {\Loc} {sit.\Pfv} {\Cop.\Prs} \\
\glt `The man is seated on the roof of the house.'

\ex \p*{merī}{kā} āṁkheṁ (\psst{Whole})

\ex \gll {aṁkh} \p*{ke}{kā} kone se\textsubscript{\rf{Locus}{Source}} (\psst{Whole}) \\
{eye} {\Gen} {corner.\Obl} {\Ins} \\
\glt `from the corner of my eye'

\ex \begin{xlist}
\ex \gll kamre \p*{ke}{kā} darvāze (\psst{Whole})\\
         {room.\Obl} {\Gen} doors\\
    \glt `the room's doors'
\ex kamre \p{meṁ} darvāze haiṁ. (\rf{Whole}{Locus})
\end{xlist}
\end{exe}

\paragraph{Sets.} Both \p{meṁ\_se} (lit. `{\Loc} {\Abl}') and \p{meṁ} (\Loc) are used to denote sets that form a \psst{Whole}. \p{meṁ\_se} is construed as \rf{Whole}{Source} since it is more literally locative in nature, while \p{meṁ} is \rf{Whole}{Locus}.

\begin{exe}
\ex \gll {in} {donoṁ} \p{meṁ\_se} {pehle} {kaun} {bolegā}? (\rf{Whole}{Source}) \\
{\Third\Pl.\Obl} {both.\Obl} {\Loc.\Abl} {first} {who} {speak.\Fut} \\
\glt `Who will speak first out of them both?'

\ex \gll sāre baccoṁ \p{meṁ} sirf tumhāre bāl lāl haiṁ. (\rf{Whole}{Locus})\\
all {child.\Pl.\Obl} {\Loc} only {\Second\Pl.\Gen} hair red {\Cop.\Prs}\\
\glt `Out of all the kids only you have red hair.'
\end{exe}

\paragraph{`Between'.} The \p{ke\_bīc} `between, among' postposition combines two or more entities into one argument to a verb (\citealt{en}):

\begin{exe}
    \ex \gll {laṛaī} {in} {donoṁ} \p{ke\_bīc} {hai} (\rf{Agent}{Whole}) \\
    {fight} {\Third\Sg.\Obl} {both.\Obl} {between} {\Cop.\Prs}\\
    \glt `(The) fight is between these two.'
\end{exe}


    

\hierCdef{Org}
\psst{Org} is not associated in the capacity of a lexical function with any marker, and is indicated by a variety of postpositions and case markers.

\begin{exe}
    \ex \gll {amerikā} {ke} {saṁyukt} {rājya} \p{ke} {rā\d sṭr\u apati} (\rf{Org}{Gestalt}) \\
    {America} {\Gen} {United} {States} {\Gen} {President} \\
    \glt `the President of the United States of America'
    
    \ex \gll {vah} {choṭe} {axbār} \p{meṁ} {kām} {kartā} {hai} (\rf{Org}{Locus}) \\
    {3SG} {small} {newspaper} {\Loc} {work} {do.\Hab} {\Cop.\Prs} \\
    \glt `He works at a small newspaper.'
    
    \ex \gll {Gūgal} \p{ke\_dvāra\_se} {āpko} {pradat} {koī} {salāh} {ya} {jānkārī} {koī} {vārantī} {nahin} {ut-pann karegī} (\rf{Org}{Agent})\\
    {Google} {by} {2.\Dat} {provide.PP} {any} {advice} {or} {information} {any} {warranty} {NEG} {create.\Fut.\Sg} \\
    \glt `Any advice or information provided to you by Google will not create any warranty.' 
    
    \ex {Gūgal} \p{ke\_sāth} {āpkā} {sambandh} {sṭeṭ} {\u af} {Kailiforniyā} {ke} {qānūn} {dvāra} {saṁcālit} {hogā}. (\rf{Org}{Ancillary})
    
    \ex {maiṁ} {sarkār} \p{ke\_liye} {kām} {kartā} {hūṁ} (\rf{Org}{Beneficiary})
\end{exe}

\hierCdef{QuantityItem}

Measure and count words (including numerals, ordinals, ordinal + measure, and numeral + measure word combinations) in Hindi largely modify the noun phrase directly, without an intervening postposition \citep{koul}. The usual marker for \psst{QuantityItem} is \p{kā}, but it is uncommon.

\begin{exe}
    \ex \gll davāoṁ \p*{kī}{kā} kamī\\
             {medicine.\Pl.\Obl} {\Gen} lack\\
        \glt `a lack of medicines'
    \ex \gll ām \p{kā} ek kilo\\
             {mango} {\Gen} one kilogram\\
        \glt `one kilogram of mangoes'
    \ex \gll seb \p{kā} ādhā (\rf{QuantityItem}{Whole})\\
             {apple} {\Gen} half\\
        \glt `one-half of the apple'
\end{exe}


    

\paragraph{Collective nouns.} The treatment of collective noun governors and their governees, follows that of \cite{en} and is labeled \rf{QuantityItem}{Stuff}:

\begin{exe}
\ex \gll {tāṛ} {ke} {vŕ\d ksoṁ} \p{kā} {jhuṇ\d d}  \\
{palm} {\Gen} {tree.\Pl} {\Gen} {grove} \\
\glt `a grove of palm trees'

\ex \gll {pilāzā} {par} {logoṁ} \p*{kī}{kā} {bhīṛ} {ikaṭṭhī} {huī} {thī} \\
{plaza} {\Loc} {person.\Pl.\Obl} {\Gen} {crowd} {assembled} {\Cop.\Pfv} {\Cop.\Pst} \\
\glt `A crowd of people had assembled at the plaza.'

\end{exe}

\hierBdef{Characteristic}
\psst{Characteristic} is expressed through \p{kā} and \p{vālā}. The difference between the two is the \p{vālā} tends to emphasise that its object is only one property (of many) of the governor.
While \p{vālā} is not a standard postposition, it mediates between nouns and noun-phrases, assigning one as a \psst{Characteristic} of the other.

\begin{exe}
    
    \ex \gll us tarah \p{kā} kām\\
             {\Third\Sg.\Obl} type {\Gen} work\\
        \glt `that kind of work'
    \ex \gll {ājkal}  \p*{kī}{kā} {duniyā} {meṁ} {log} {aise} {hote} hain (\rf{Time}{Characteristic})\\
    {today} {\Gen} {world} {\Loc} {people} {COMP.\Pl} {exist.\Hab.\Pl} {\Cop.\Pl} \\
    \glt `People are like that in today's world.'
    \ex \begin{xlist}
    \ex \gll {Jāpān} {duniyā} {ke} {tīsrā} {sabse baṛā} {tel} {khapat} \p{vālā} {deś} {hai} \\
     {Japan} {world} {\Gen} {third} {largest} {oil} {consumption} {\Adj} {country} {\Cop.\Sg} \\
     \glt `Japan is the third largest oil-consuming country of the world.'
    \ex \gll nīlā \p{vālā} ghar\\
             {blue} {\Adj} house\\
        \glt `a house that is blue'
    \ex {do} {sāl} {kī\textsubscript{\psst{Identity}}} {umr} \p{vālā} {kuttā}
    \ex ūpar \p{vālā} kamrā
    (\rf{Locus}{Characteristic})
    \ex pīne \p{vālā} sāf pāni (\rf{Purpose}{Characteristic})
    \end{xlist}
    \ex \gll rāy \p{meṁ} fark\\
             opinion {\Loc} difference\\
        \glt `a difference in opinion'
    \ex umr hogī gyārah sāl lekin lambāī \p{meṁ} zyādā baṛā lagtā hai.
    \ex khilāṛī vazan \p{ke\_hisāb\_se} cune gaye.
\end{exe}

\paragraph{Containers.} Like in English, \psst{Characteristic} construed as \psst{Stuff} described containers that are filled with something.

\begin{exe}
    
    \ex \gll {pānī} \p*{kī}{kā} {botal} {20} {rūpye} {kī} {hai}. (\rf{Characteristic}{Stuff}) \\
    {water} {\Gen} {bottle} {20} {rupees} {\Gen} {\Cop.\Sg} \\
    \glt `The bottle of water costs 20 rupees.'

    \ex {tumhāre} {gahne} {aur} {kapṛoṁ} \p{kā} {baksā} (\rf{Characteristic}{Stuff})
\end{exe}

\paragraph{Examining for an attribute.} \p{ke\_liye} is used in transitive verb contexts where the attribute of the \psst{Theme} is being examined.

\begin{exe}
    \ex \gll bacce ne rākṣasoṁ \p{ke\_liye} kamre kī\textsubscript{\psst{Theme}} jāṁc kī.\\
             {child.\Obl} {\Erg} {demon.\Pl.\Obl} {for} {room} {\Gen} {checking} {do.\Pfv}\\
        \glt `The child checked the room for monsters.'
\end{exe}


\paragraph{States.} The state or condition that an entity is in is \rf{Characteristic}{Locus}.

\begin{exe}
 
    \ex \gll kitāb Pañjābī \p{meṁ} hai. \\
             {book} {Punjabi} {\Loc} {\Cop.\Prs}\\
        \glt `The book is in Punjabi.'
    \ex vah kis hālat \p{meṁ} hai?
    \ex acambhe \p{meṁ}
    \ex trikoṇ \p{ke\_rūp\_meṁ}
\end{exe}


\hierCdef{Possession}
The genitive \p{kā} and adjectival \p{vālā} indicate a \psst{Possession} when its object is the item being possessed and the governor is a possessor (i.e. the reverse of the genitive \psst{Possessor}).

\begin{exe}
    \ex \gll {vah} {ghar} \p{kā} {mālik} {hai}. \\
    {\Third\Sg} {house} {\Gen} {owner} {\Cop.\Prs} \\
    \glt `He is the owner of the house.'
    \ex \gll {vah} {kāfī} {paise} \p{vālā} {thā}.  (\rf{Possession}{Characteristic})   \\
 {\Third\Sg} {quite} {money} {\Adj} {\Cop.\Pst}  \\
 \glt `He was quite rich.'
    \ex \gll binā paise \p{kā} ādmī\\
             without money {\Gen} man\\
        \glt `a man without money'
\end{exe}

\paragraph{Verbal arguments.} The morphosyntactic \psst{Theme} argument to a verb (marked with an accusative-type postposition or \p{ko}) dealing with change of possession or transfer of goods and services is labelled \rf{Possession}{Theme}.

\begin{exe}
    \ex \gll {unhone} {pākśāstr\u a} {kī} {kitāboṁ} \p{par} {khūb} {xarc} {kiyā} {hai}  \\
    {\Third\Pl.\Erg} {cooking} {\Gen} {book.\Pl.\Obl} {\Loc} {lot} {spend} {do.\Pfv}  {\Cop.\Prs} \\
    \glt `He has spent a lot on cookbooks.'
    \ex maiṁ us khilaune \p{ko} xarīdnā cāhtā hūṁ!
\end{exe}



    

\hierCdef{PartPortion}

\begin{exe}
    \ex \gll naī iñjan \p*{vālī}{vālā} gāṛī (\rf{PartPortion}{Characteristic})\\
             {new} {engine} {\Adj} {car}\\
        \glt `a car with a new engine'
    \ex \gll do darvāzoṁ \p{vālā} kamrā (\rf{PartPortion}{Characteristic})\\
             two {door.\Pl.\Obl} {\Adj} room\\
        \glt `a room with two doors'
\end{exe}

\paragraph{\textit{binā} and \p{kā}/\p{vālā}.} As a postposition, \p{ke\_binā} can mark an NP as \psst{PartPortion}, indicating an \texttt{obl} argument to a verb.

\begin{exe}
    \ex \gll {masālā} \p{ke\_binā} {pūrī-masālā} {kyā} {hai}? \\
    {spices} {without} {puri-spices} {what} {\Cop.\Sg} \\
    \glt `What is puri-spices without the spices?'
    
    \ex \gll {rāhul} {drāvid} \p{ke\_binā} {kyā} {hotā} {hai} {bhāratīya} {ballebāzī} {kā} {hāl}? \\
    {Rahul} {Dravid} {without} {what} {be.\Prs} {\Cop.\Sg} {indian} {batting} {\Gen} {state} \\
    \glt `What is the state of Indian batting without Rahul Dravid?'
\end{exe}

As a noun modifier, \textit{binā} is often coordinated with the postpositions \p{kā} or \p{vālā}. In these cases, we do not label \textit{binā}, but we label the coordinating postposition \psst{PartPortion}. The reasoning is that when \p{binā} is dropped, the coordinating postpositions still provide the same semantics (e.g.~\textit{cīnī vālī cāy} 'tea with sugar').

\begin{exe}
    \ex \gll binā cīnī \p*{vālī}{vālā}\textsubscript{\rf{PartPortion}{Characteristic}} cāy\\
             without sugar {\Adj} tea\\
        \glt `tea without sugar'
    \ex binā cīnī \p{kā}\textsubscript{\psst{PartPortion}} dūdh
\end{exe}

\paragraph{Sets.} Non-members and members of a set can be marked \psst{PartPortion} by \p{ke\_alāvā} `besides, other than' and \p{ke\_atirikt} (`in addition to').

\begin{exe}

    \ex \gll {śahr} \p{ke\_alāvā} {gāvoṁ} {meṁ} {bhī} {gas} {kanekśan} {baṛh} {rahe} {hai} \\
    {city} {other.than} {village.\Pl.\Obl} {\Loc} {too} {gas} {connection} {increase} {\Cont} {\Cop.\Prs} \\
    \glt `Other than in the city, gas installations are increasing in the villages too.' 
    \ex {Smith} {aur} {Kailis} \p{ke\_alāvā}
    \ex {Latā} {Mangeśkar}, {Āśā} {Bhosle} {to} {niyamit} {āvāze} {thīṁ} {hī}, \p*{inke\_atirikt}{ke\_atirikt} {Hemant} {Kumār}, {Talat} {Mahmūd} {bhī}
\end{exe}

\p{jaisā} `such as' can also mark set members.

\begin{exe}
    \ex \gll {Dīwālī} {aur} {Holī} \p*{jaise}{jaisā} {bhāratīya} {tyauhār} {manātīṁ} {haiṁ} \\
    {diwali} {and} {holi} {like} {indian} {festival} {celebrate.\Hab} {\Cop.\Prs} \\
    \glt `They celebrate Indian festivals like diwali and holi.'
    
    \ex {h\u okī}, {fuṭb\u ol}, {aur} {krikeṭ} \p*{jaise}{jaisā} {pāramparik} {khel}
    
\end{exe}

\hierDdef{Stuff}

\psst{Stuff} is marked by the genitive \p{kā}, and it is not different from how the English guidelines treat it.

\begin{exe}
    \ex \gll sone \p*{kī}{kā} thālī (\psst{Stuff})\\
             {gold.\Obl} {\Gen} {platter}\\
        \glt `a platter made of gold'
    \ex \gll {bīyar} \p*{kī}{kā} {botal} (\rf{Characteristic}{Stuff}) \\
        {beer} {\Gen} {bottle} \\
        \glt `bottle of beer'
    \ex {vṛ\d ksoṁ} \p{kā} {jhun\d d} (\rf{QuantityItem}{Stuff})
    \ex {logoṁ} \p*{kī}{kā} {bhīṛ} (\rf{QuantityItem}{Stuff})
    \ex {chātroṁ} \p*{kī}{kā} {kak\d sā} (\rf{OrgMember}{Stuff})
    \ex {cricket} {vāloṁ} \p*{kī}{kā} {tīm} (\rf{OrgMember}{Stuff})
\end{exe}

\hierCdef{OrgMember}

\psst{OrgMember} is largely marked by the the genitive \p{kā}.

\begin{exe}
    \ex \gll {mere} {beṭe} \p{kā} {parivār} (\rf{OrgMember}{Gestalt}) \\
    {1\Sg.\Gen} {child.\Obl} {\Gen} {family} \\
    \glt `My child's family.'
    
    \ex \gll {coroṁ} \p{kī} {dhāṛ} (\rf{OrgMember}{Stuff}) \\
    {thief.\Pl.\Obl} {\Gen} {gang} \\
    \glt `gang of thieves'
    
    \ex \gll \p*{merī}{kā} {kampanī} (\rf{OrgMember}{Possessor}) \\
    {1\Sg.\Gen} {company} \\
    \glt `My company.'
    
\end{exe}

\hierCdef{QuantityValue}

\psst{QuantityValue} is uncommon, but is indicated by genitive \p{kā}.

\begin{exe}
    \ex \gll ek kilo \p{kā} ām\\
             one {kilogram} {\Gen} mango\\
        \glt `a one-kilogram mango'
\end{exe}

\hierDdef{Approximator}

\psst{Approximator} is indicated by a number of targets, all dealing with scalar comparisons.

\begin{exe}
    \ex \p{lagbhag} `around, approximately':\begin{xlist}
    \ex \gll {unkī} {kampanī} {Dillī} {ke} \p{lagbhag} {ek karoṛ} garīboṁ tak bijlī pahuṁchātī hai\\
     {\Third\Sg.\Gen} {company} {Delhi} {\Gen} around {one crore} {pauper.\Pl.\Obl} to electricity {deliver.\Prs} {\Cop.\Sg}   \\
    \glt `His company provides electricity to around 10 million of Delhi's poor'
    
    \ex {gāv} {ke} \p{lagbhag} chār sau log
    \end{xlist}
    
    \ex \p{qarīb} `nearly, almost':
    \begin{xlist}
    \ex \gll {Eyar} {Inḍīyā} {ke} \p{qarīb} {ādhe} {pāylaṭoṁ} kī {haṛtāl}\\
     Air India {\Gen} {nearly} {half} {pilot.\Pl.\Obl} {\Gen} {strike}  \\
    \glt `the strike of nearly half of Air India's pilots'
    
    \ex {uṛān bharne} {ke} \p{qarīb} {20} {minat} {bād}
    \end{xlist}
    \ex \p{ke\_adhik} / \p{se\_adhik}, \p{se\_zyādā} `over, greater than':
    \begin{xlist}
        \ex \gll 70 {fīsadī} \p{ke\_adhik}\\
                 {70} {percent} {greater.than}\\
            \glt `greater than 70 percent'
        \ex \gll {jodhpūr} {ke} 1200 \p{se\_adhik} {d\u aktar} {haṛtal} par {the} \\
        {jodhpur} {\Gen} 1200 {more than} doctor strike {\Loc} {\Cop.\Pst}  \\
        \glt `More than 1200 doctors from Jodhpur were on strike'
        \ex Inglaind ne vah ṭesṭ 300 \p{ke\_adhik} {antar}  se jītā 
    \end{xlist}
    \ex \p{ke\_bīc} `between':
    \begin{xlist}
    \ex \gll {ummīd} {hai} ki hum {pānc} se {cha\d h} hazār \p{ke\_bīc} nayī {logoṁ} kī bhartī {kareṅge} \\
    hope {\Cop.\Sg} COMP {3.\Pl} five {\Com} six thousand between new {person.\Pl} {\Gen} recruit {do.\Pl.\Fut} \\
    \glt `The hope is that we will recruit between five to six thousand new recruits'
    \end{xlist}
    \ex \p{ke\_āspās} `close to, near':\begin{xlist}
    \ex \gll {d\u alar} {ke mukābale} {rūpyā} 52 {rūpye} \p{ke\_āspās} {pahu\d ncā} \\
    dollar {compared to} rupee 52 rupees {close to} reach.\Pfv   \\
    \glt `The Rupee reached close to 52 rupees (compared) to the Dollar'
    
    \ex \gll {gyārah} {baje} \p{ke\_āspās} vah {dillī} {pahunchī} \\
    eleven time {close to} {3.\Sg} Delhi {reach.\Pfv} \\
    \glt `He reached Delhi close to eleven o'clock'
    \end{xlist}
\end{exe}

\paragraph{Confusion with \rf{ComparisonRef}{Locus}.} The difference between \psst{Approximator} and the more literal \rf{ComparisonRef}{Locus} can be clearly defined by syntax, although it does have a semantic element.

When the number marked by the approximating postposition is a predicate with the copula, than it is \rf{ComparisonRef}{Locus} (since it is comparing an unknown value to a point or points on a scale). If it is modifying an NP, then it is \psst{Approximator}.

Some examples of \rf{ComparisonRef}{Locus} follow.

\begin{exe}
    \ex \gll iskī qīmat pāṁc se sāt lākh \p{ke\_bīc} hai  \\
    {\Third\Sg.\Gen} cost five {\Abl} {seven} {lakh} between \Cop.\Sg   \\
    \glt `The cost of this is between five to seven lakhs (five to seven hundred thousand).'
    
    \ex \gll {ūṁchāī} 5 {mīṭar} \p{se\_kam} {nahiṁ} honī {cāhiye} \\
    height 5 metre {less.than} {\Neg} {\Cop.\Inf} {ought} \\
    \glt `(The) height should not be less than five metres.'

\end{exe}

\hierBdef{Ensemble}

\psst{Ensemble} by itself is rare in Hindi, rather expressed through compounding (two adjacent words in one NP) or a conjunction such as \textit{aur} `and'.

Verb arguments that are inanimate and marked with a postposition or case marker similar to English \textit{with} are \rf{Ensemble}{Ancillary}.

\begin{exe}
     \ex mujhe cāval \p{ke\_sāth} dāl cāhiye. (\rf{Ensemble}{Ancillary})
\end{exe}

If, however, these can be better interpreted as one whole NP (with the postposition-marked term being a UD \texttt{nmod} to the head), then plain \psst{Ensemble} applies.

\hierBdef{ComparisonRef}

\psst{ComparisonRef} is typically marked by \p{se} (\Abl) `than', \p{jaisā} / \p{ke\_jaisā} ('like', comparing NPs), and \p{jaise} / \p{ke\_jaise} (`like', adverbial). The latter two are also equivalent to \p{kī\_tarah} and \p{kī\_bhāṁti}.

\begin{exe}
    \ex \gll dahī cāval \p{se} acchā koī khānā {nahīṁ} hai. \\
    curd rice {\Abl} good {any} food {\Neg} {\Cop.\Prs} \\
    \glt `There is no food as good as curd--rice.'
    
    \ex \gll ek citra hazār {śabdoṁ} \p{se} {bahtar} hai. \\
    one picture thousand {word.\Pl.\Obl} {\Abl} beter {\Cop.\Sg} \\
    \glt`A picture is better than a thousand words.'
    
    \ex mujh \p{jaisā} ādmī
    \ex \p*{mere\_jaisā}{ke\_jaisā} ādmī
    \ex \gll \p*{uskī\_jagah}{kī\_jagah} yah cāhiye.\\
             {\Third\Sg.in.place.of} {\Third\Sg} {wanted}\\
        \glt `I want this instead of that.'
\end{exe}

\paragraph{Sufficiency/excess.} \p{ke\_liye} handles sufficiency/excess comparisons, and is labelled \rf{ComparisonRef}{Purpose} in such a usage \citep[60]{FortuinExcess}.

\begin{exe}
    \ex skūl jāne \p{ke\_liye} vah kāfī baṛā hai (\rf{ComparisonRef}{Purpose})
\end{exe}

\paragraph{Adverbial.} The adverbial \p{jaise} / \p{ke\_jaise} can be read as either indicating an analogy (\rf{Manner}{ComparisonRef}) or a conclusion (\rf{Theme}{ComparisonRef}). The latter reading is especially likely for experiencer verbs (e.g.~\textit{lagnā} `to seem'), in which case one can try paraphrasing with a complementiser: \textit{lagtā hai ki...}. If the paraphrase works, then the conclusion reading is more salient.

\begin{exe}
    \ex \rf{Theme}{ComparisonRef}:\begin{xlist}
        \ex \gll aisā lagā \p{jaise} vah jhūṭh bol rahā hai.\\
                 {like.this} {feel.\Pfv} {like} {\Third\Sg} lie say {\Cont} {\Cop.\Prs}\\
            \glt `It seemed like he was lying.'
        \ex lagā ki vah jhūṭh bol rahā hai.
    \end{xlist}
    \ex \rf{Manner}{ComparisonRef}:\begin{xlist}
        \ex \gll aisā lagā \p{jaise} pūre deś kā khānā khā liyā hai.\\
                 {like.this} {feel.\Pfv} {like} {whole.\Obl} {country} {\Gen} {food} {eat} {take.\Pfv} {\Cop.\Prs} \\
            \glt `It felt like I ate the whole country's food supply.'
        \ex \#lagā ki pūre deś kā khānā khā liyā hai.
    \end{xlist}
\end{exe}

\paragraph{Implicit comparison.} Implicit comparison (instead of a direct comparison of an attribute) is also indicated \psst{ComparisonRef} \citep{zyaadaa}.

\begin{exe}
    \ex \gll us nibandh \p{ke\_muqāble} ye nibandh lambā hai.\\
             {\Third\Sg} essay {against} {\Third\Sg} essay long {\Cop.\Prs}\\
        \glt `In comparison to that essay, this essay is longer'
    \ex zindagi \p{ke\_banisbat} ġulāmī pyārī hai?
\end{exe}

\hierBdef{RateUnit}

This is rare in Hindi, and is only directly expressed by the high-register \p{prati} (in Hindi, a Sanskrit borrowing) and \p{fī} (in Urdu, a Perso-Arabic borrowing).

\begin{exe}
    \ex \p{prati} vyakti
    \ex \p{fī} śaxs
\end{exe}

\hierBdef{SocialRel}

The genitive \p{kā} marks \rf{SocialRel}{Gestalt}. Note also that some verbs (e.g.~\textit{dostī karnā} `to befriend') license their arguments as \psst{SocialRel}.

\begin{exe}
    \ex \gll {ā} {gayā} \p*{terā}{kā} bhāī\\
    {come} {go.\Pfv} {2\Sg.\Gen} {brother}\\
    \glt `Your brother has come.'
    \ex pikcar abhī bākī hai \p*{mere}{kā} dost!
    \ex \p*{merī}{kā} jān sabse pyāri hai.
\end{exe}

\hierAdef{Context}

\hierBdef{Focus}

The traditional emphatic particles (\p{hī} `only', \p{bhī} `also', \p{to} contrastive, and some uses of \p{tak} `even') are all labelled \psst{Focus}. They are postposition-like, in that they place emphasis on the preceding element in relation to its governor.

\begin{exe}
    \ex maiṁ \p{hī} ghar jāūṅgā.
    \ex tū \p{to} ghar nahīṁ jāegā.
    \ex Rāhul, nām \p{to} sunā hī hogā.
\end{exe}

\section{Special labels}

\subsection{DISCOURSE (\olbl{\textasciigrave d})}

When uses quotatively, \p{ko} and \p{ke\_liye} are labelled \olbl{\textasciigrave d}. These are equivalent to the English infinitival \textit{to}, hence we agree with the labelling of \citet{en}.

\begin{exe}
    \ex \gll usne jāne \p{ko} kahā.\\
             {\Third\Sg.\Erg} {go.\Inf.\Obl} {\Dat} {say.\Pfv}\\
        \glt `He said to go.'
    \ex vah tumse bāt karne \p{ke\_liye} bolā.
\end{exe}

\bibliographystyle{plainnat}
\bibliography{references.bib}

\begin{thebibliography}{34}
\providecommand{\natexlab}[1]{#1}
\providecommand{\url}[1]{\texttt{#1}}
\expandafter\ifx\csname urlstyle\endcsname\relax
  \providecommand{\doi}[1]{doi: #1}\else
  \providecommand{\doi}{doi: \begingroup \urlstyle{rm}\Url}\fi

\bibitem[Anwar et~al.(2016)Anwar, Bhat, Sharma, Vaidya, Palmer, and
  Khan]{anwar-etal-2016-proposition}
Maaz Anwar, Riyaz~Ahmad Bhat, Dipti Sharma, Ashwini Vaidya, Martha Palmer, and
  Tafseer~Ahmed Khan.
\newblock A {P}roposition {B}ank of {U}rdu.
\newblock In \emph{Proceedings of the Tenth International Conference on
  Language Resources and Evaluation ({LREC}'16)}, pages 2379--2386,
  Portoro{\v{z}}, Slovenia, May 2016. European Language Resources Association
  (ELRA).
\newblock URL \url{https://www.aclweb.org/anthology/L16-1377}.

\bibitem[Arora and Schneider(2020)]{arora-etal-2020-snacs}
Aryaman Arora and Nathan Schneider.
\newblock {SNACS} annotation of case markers and adpositions in {H}indi.
\newblock In \emph{Proceedings of the Second Workshop on Computational Research
  in Linguistic Typology}, Online, 2020. Association for Computational
  Linguistics.
\newblock URL \url{https://sigtyp.github.io/workshops/2020/papers/8.pdf}.

\bibitem[Arora et~al.(2021)Arora, Venkateswaran, and
  Schneider]{arora-etal-2021-snacs}
Aryaman Arora, Nitin Venkateswaran, and Nathan Schneider.
\newblock {SNACS} annotation of case markers and adpositions in {H}indi.
\newblock In \emph{Proceedings of the Society for Computation in Linguistics},
  volume~4, pages 454--458, Online, 2021. Society for Computation in
  Linguistics.
\newblock URL \url{https://scholarworks.umass.edu/scil/vol4/iss1/57/}.

\bibitem[Begum and Sharma(2010)]{begum-sharma-2010-preliminary}
Rafiya Begum and Dipti~Misra Sharma.
\newblock A preliminary work on {H}indi causatives.
\newblock In \emph{Proceedings of the Eighth Workshop on {A}sian Language
  Resouces}, pages 120--128, Beijing, China, August 2010. Coling 2010
  Organizing Committee.
\newblock URL \url{https://www.aclweb.org/anthology/W10-3216}.

\bibitem[Bhat et~al.(2014)Bhat, Jain, Vaidya, Palmer, Ahmed~Khan, Sharma, and
  Babani]{bhat-etal-2014-adapting}
Riyaz~Ahmad Bhat, Naman Jain, Ashwini Vaidya, Martha Palmer, Tafseer
  Ahmed~Khan, Dipti~Misra Sharma, and James Babani.
\newblock Adapting predicate frames for {U}rdu {P}rop{B}anking.
\newblock In \emph{Proceedings of the {EMNLP}{'}2014 Workshop on Language
  Technology for Closely Related Languages and Language Variants}, pages
  47--55, Doha, Qatar, October 2014. Association for Computational Linguistics.
\newblock \doi{10.3115/v1/W14-4206}.
\newblock URL \url{https://www.aclweb.org/anthology/W14-4206}.

\bibitem[Bhatia et~al.(2013{\natexlab{a}})Bhatia, Vaidya, Narasimhan, and
  Palmer]{propbank}
Archna Bhatia, Ashwini Vaidya, Bhuvana Narasimhan, and Martha Palmer.
\newblock Hindi {P}rop{B}ank annotation guidelines, 2013{\natexlab{a}}.
\newblock URL
  \url{http://verbs.colorado.edu/hindiurdu/guidelines_docs/PBAnnotationGuidelines.pdf}.

\bibitem[Bhatia(2016)]{bhatia2016}
Sakshi Bhatia.
\newblock Causation in {H}indi-{U}rdu: Care for your instruments and subjects.
\newblock In Rahul Balusu and Sandhya Sundaresan, editors, \emph{Proceedings of
  FASAL 5}, 2016.
\newblock URL
  \url{https://ojs.ub.uni-konstanz.de/jsal/index.php/fasal/article/view/83}.

\bibitem[Bhatia et~al.(2013{\natexlab{b}})Bhatia, Iyer, and Kaur]{zyaadaa}
Sakshi Bhatia, Jyoti Iyer, and Gurmeet Kaur.
\newblock Comparatives in {H}indi-{U}rdu: Puzzling over {ZYAADAA}.
\newblock \emph{LISSIM Working Papers}, 1\penalty0 (1):\penalty0 15--28,
  2013{\natexlab{b}}.
\newblock URL
  \url{https://blogs.umass.edu/jiyer/files/2018/04/BHATIA-IYER-KAUR_2013_LWP.pdf}.

\bibitem[Bhatt et~al.(2013)Bhatt, Farudi, and Rambow]{psg}
Rajesh Bhatt, Annahita Farudi, and Owen Rambow.
\newblock {H}indi-{U}rdu phrase structure annotation guidelines, 2013.
\newblock URL
  \url{http://verbs.colorado.edu/hindiurdu/guidelines_docs/PhraseStructureguidelines.pdf}.

\bibitem[Butt(1993)]{buttthesis}
Miriam Butt.
\newblock \emph{The Structure of Complex Predicates in {U}rdu}.
\newblock PhD thesis, Stanford University, 1993.

\bibitem[Butt et~al.(2006)Butt, Grimm, and Ahmed]{dativesubjects}
Miriam Butt, Scott Grimm, and Tafseer Ahmed.
\newblock Dative subjects.
\newblock In \emph{NWO/DFG Workshop on Optimal Sentence Processing}, 2006.
\newblock URL
  \url{https://ling.sprachwiss.uni-konstanz.de/pages/home/butt/main/papers/nijmegen-hnd.pdf}.

\bibitem[{de Hoop} and Narasimhan(2005)]{dehoop2005dam}
Helen {de Hoop} and Bhuvana Narasimhan.
\newblock Differential case-marking in {H}indi.
\newblock In Mengistu Amberber and Helen {De Hoop}, editors, \emph{Competition
  and Variation in Natural Languages}, Perspectives on Cognitive Science, pages
  321--345. Elsevier, Oxford, 2005.
\newblock \doi{https://doi.org/10.1016/B978-008044651-6/50015-X}.
\newblock URL
  \url{http://www.sciencedirect.com/science/article/pii/B978008044651650015X}.

\bibitem[Fortuin(2013)]{FortuinExcess}
Egbert Fortuin.
\newblock The construction of excess and sufficiency from a crosslinguistic
  perspective.
\newblock \emph{Linguistic Typology}, 17\penalty0 (1):\penalty0 31--88, 2013.
\newblock \doi{doi:10.1515/lity-2013-0002}.
\newblock URL \url{https://doi.org/10.1515/lity-2013-0002}.

\bibitem[Goel et~al.(2020)Goel, Prabhu, Debnath, Modi, and
  Shrivastava]{goel-etal-2020-hindi}
Pranav Goel, Suhan Prabhu, Alok Debnath, Priyank Modi, and Manish Shrivastava.
\newblock {H}indi {T}ime{B}ank: An {ISO}-{T}ime{ML} annotated reference corpus.
\newblock In \emph{16th Joint ACL - ISO Workshop on Interoperable Semantic
  Annotation PROCEEDINGS}, pages 13--21, Marseille, May 2020. European Language
  Resources Association.
\newblock ISBN 979-10-95546-48-1.
\newblock URL \url{https://www.aclweb.org/anthology/2020.isa-1.2}.

\bibitem[Hautli-Janisz et~al.(2015)Hautli-Janisz, King, and
  Ramchand]{hautli-janisz-etal-2015-encoding}
Annette Hautli-Janisz, Tracy~Holloway King, and Gilian Ramchand.
\newblock Encoding event structure in {U}rdu/{H}indi {V}erb{N}et.
\newblock In \emph{Proceedings of the The 3rd Workshop on {EVENTS}: Definition,
  Detection, Coreference, and Representation}, pages 25--33, Denver, Colorado,
  June 2015. Association for Computational Linguistics.
\newblock \doi{10.3115/v1/W15-0804}.
\newblock URL \url{https://www.aclweb.org/anthology/W15-0804}.

\bibitem[Hwang et~al.(2020)Hwang, Choe, Han, and Schneider]{hwang-etal-2020-k}
Jena~D. Hwang, Hanwool Choe, Na-Rae Han, and Nathan Schneider.
\newblock K-{SNACS}: Annotating {K}orean adposition semantics.
\newblock In \emph{Proceedings of the Second International Workshop on
  Designing Meaning Representations}, pages 53--66, Barcelona Spain (online),
  December 2020. Association for Computational Linguistics.
\newblock URL \url{https://www.aclweb.org/anthology/2020.dmr-1.6}.

\bibitem[Hwang et~al.(2021)Hwang, Han, Choe, and Schneider]{korean}
Jena~D. Hwang, Na-Rae Han, Hanwool Choe, and Nathan Schneider.
\newblock Korean adposition and case supersenses v0.9.
\newblock Unpublished, 2021.

\bibitem[Kachru(1970)]{kachru1970}
Yamuna Kachru.
\newblock A note on possessive constructions in {H}indi-{U}rdu.
\newblock \emph{Journal of Linguistics}, 6\penalty0 (1), 1970.
\newblock URL \url{https://www.jstor.org/stable/4175050}.

\bibitem[Kachru(2009)]{hindiurdu}
Yamuna Kachru.
\newblock {H}indi--{U}rdu.
\newblock In Bernard Comrie, editor, \emph{The World's Major Languages}, pages
  399--416. Routledge, 2 edition, 2009.

\bibitem[Khan(2009)]{tafseer}
Tafseer~Ahmed Khan.
\newblock \emph{Spatial Expressions and Case in {S}outh {A}sian Languages}.
\newblock PhD thesis, University of Konstanz, 2009.
\newblock URL \url{http://kops.uni-konstanz.de/handle/123456789/12508}.

\bibitem[Koul(2008)]{koul}
Omkar~N. Koul.
\newblock \emph{Modern {H}indi {G}rammar}.
\newblock Dunwoody Press, 2008.
\newblock ISBN 978-1-931546-06-5.
\newblock URL \url{http://www.koausa.org/iils/pdf/ModernHindiGrammar.pdf}.

\bibitem[Masica(1993)]{masica1993indo}
Colin~P. Masica.
\newblock \emph{The {I}ndo-{A}ryan {L}anguages}.
\newblock Cambridge University Press, 1993.

\bibitem[Mohanan and Verma(1990)]{experiencersubjects}
K.~P. Mohanan and Mahendra~K. Verma, editors.
\newblock \emph{Experiencer Subjects in South Asian Languages}.
\newblock Cambridge University Press, 1990.

\bibitem[Mohanan(1994)]{mohanan1994argument}
Tara Mohanan.
\newblock \emph{Argument structure in {H}indi}.
\newblock {C}enter for the {S}tudy of {L}anguage ({CSLI}), 1994.

\bibitem[Narasimhan(2003)]{narasimhan2003motion}
Bhuvana Narasimhan.
\newblock Motion events and the lexicon: a case study of {H}indi.
\newblock \emph{Lingua}, 113\penalty0 (2):\penalty0 123--160, 2003.

\bibitem[Nivre et~al.(2020)Nivre, de~Marneffe, Ginter, Hajič, Manning,
  Pyysalo, Schuster, Tyers, and Zeman]{nivre2020universal}
Joakim Nivre, Marie-Catherine de~Marneffe, Filip Ginter, Jan Hajič,
  Christopher~D. Manning, Sampo Pyysalo, Sebastian Schuster, Francis Tyers, and
  Daniel Zeman.
\newblock Universal dependencies v2: An evergrowing multilingual treebank
  collection, 2020.

\bibitem[Palmer et~al.(2009)Palmer, Bhatt, Narasimhan, Rambow, Sharma, and
  Xia]{hindisyntax}
Martha Palmer, Rajesh Bhatt, Bhuvana Narasimhan, Owen Rambow, Dipti~Misra
  Sharma, and Fei Xia.
\newblock Hindi syntax: Annotating dependency, lexical predicate-argument
  structure, and phrase structure.
\newblock In \emph{Proceedings of the 7th International Conference on Natural
  Language Processing}, 2009.
\newblock URL
  \url{http://faculty.washington.edu/fxia/mpapers/2009/ICON2009.pdf}.

\bibitem[Prange and Schneider(2021)]{german}
Jakob Prange and Nathan Schneider.
\newblock Draw mir a sheep: A supersense-based analysis of {G}erman case and
  adposition semantics.
\newblock \emph{Künstliche Intelligenz}, 35\penalty0 (2), 2021.

\bibitem[Ramchand(2011)]{se-causative}
Gillian~Catriona Ramchand.
\newblock Licensing of instrumental case in {H}indi/{U}rdu causatives.
\newblock \emph{Nordlyd}, 38:\penalty0 49--71, 2011.

\bibitem[Schneider et~al.(2018)Schneider, Hwang, Srikumar, Prange, Blodgett,
  Moeller, Stern, Bitan, and Abend]{schneider-18}
Nathan Schneider, Jena~D. Hwang, Vivek Srikumar, Jakob Prange, Austin Blodgett,
  Sarah~R. Moeller, Aviram Stern, Adi Bitan, and Omri Abend.
\newblock Comprehensive supersense disambiguation of {E}nglish prepositions and
  possessives.
\newblock In \emph{Proc. of {ACL}}, pages 185--196, Melbourne, Australia, July
  2018.

\bibitem[Schneider et~al.(2020)Schneider, Hwang, Bhatia, Han, Srikumar,
  O'Gorman, and Abend]{en}
Nathan Schneider, Jena~D. Hwang, Archna Bhatia, Na{-}Rae Han, Vivek Srikumar,
  Tim O'Gorman, and Omri Abend.
\newblock Adposition and case supersenses v2.5: Guidelines for english.
\newblock \emph{CoRR}, abs/1704.02134, 2020.
\newblock URL \url{http://arxiv.org/abs/1704.02134}.

\bibitem[Shalev et~al.(2019)Shalev, Hwang, Schneider, Srikumar, Abend, and
  Rappoport]{shalev-19}
Adi Shalev, Jena~D. Hwang, Nathan Schneider, Vivek Srikumar, Omri Abend, and
  Ari Rappoport.
\newblock Preparing {SNACS} for subjects and objects.
\newblock In \emph{Proc. of the First International Workshop on Designing
  Meaning Representations}, pages 141--147, Florence, Italy, August 2019.

\bibitem[Vaidya et~al.(2011)Vaidya, Choi, Palmer, and
  Narasimhan]{vaidya-etal-2011-analysis}
Ashwini Vaidya, Jinho Choi, Martha Palmer, and Bhuvana Narasimhan.
\newblock Analysis of the {H}indi {P}roposition {B}ank using dependency
  structure.
\newblock In \emph{Proceedings of the 5th Linguistic Annotation Workshop},
  pages 21--29, Portland, Oregon, USA, June 2011. Association for Computational
  Linguistics.
\newblock URL \url{https://www.aclweb.org/anthology/W11-0403}.

\bibitem[Vaidya et~al.(2013)Vaidya, Palmer, and
  Narasimhan]{vaidya-etal-2013-semantic}
Ashwini Vaidya, Martha Palmer, and Bhuvana Narasimhan.
\newblock Semantic roles for nominal predicates: Building a lexical resource.
\newblock In \emph{Proceedings of the 9th Workshop on Multiword Expressions},
  pages 126--131, Atlanta, Georgia, USA, June 2013. Association for
  Computational Linguistics.
\newblock URL \url{https://www.aclweb.org/anthology/W13-1018}.

\end{thebibliography}


\printindex
\printindex[construals]
\printindex[revconstruals]

\end{document}